\documentclass[sigconf]{acmart}
\usepackage{bbm}
\usepackage[ruled,linesnumbered]{algorithm2e}
\usepackage{multirow}
\usepackage{algpseudocode}
\usepackage{enumitem}
\usepackage{graphicx}
\usepackage{subfigure}
\usepackage{bm}
\usepackage{hyperref}

\newcommand{\bluefont}[1]{ {\color{blue}{#1}}}

\AtBeginDocument{%
  \providecommand\BibTeX{{%
    \normalfont B\kern-0.5em{\scshape i\kern-0.25em b}\kern-0.8em\TeX}}}

\copyrightyear{2022}
\acmYear{2022}
\setcopyright{rightsretained} 

\acmConference[KDD '22]{Proceedings of the 28th ACM SIGKDD Conference on Knowledge Discovery and Data Mining}{August 14--18, 2022}{Washington, DC, USA} 
\acmBooktitle{Proceedings of the 28th ACM SIGKDD Conference on Knowledge Discovery and Data Mining (KDD '22), August 14--18, 2022, Washington, DC, USA}
\acmISBN{978-1-4503-9385-0/22/08}
\acmDOI{10.1145/3534678.3539484}

\usepackage{etoolbox}
\makeatletter
\patchcmd{\maketitle}{\@copyrightpermission}{
   \begin{minipage}{0.3\columnwidth}
     \href{https://creativecommons.org/licenses/by-nc-sa/4.0/}{\includegraphics[width=0.90\textwidth]{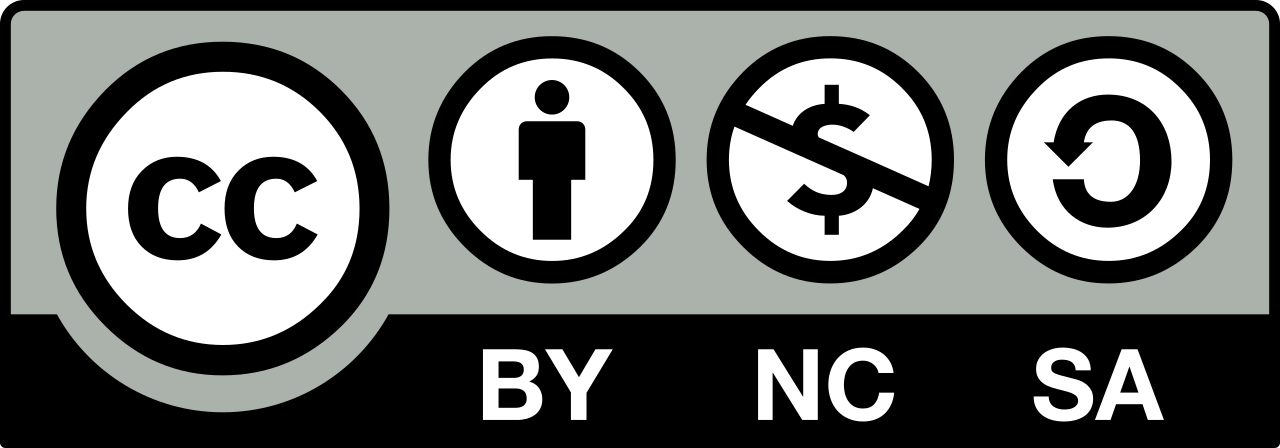}}
   \end{minipage}\hfill
   \begin{minipage}{0.7\columnwidth}
     \href{hhttps://creativecommons.org/licenses/by-nc-sa/4.0/}{This work is licensed under a Creative Commons Attribution-NonCommercial-ShareAlike International 4.0 License.}
   \end{minipage}
  
   \vspace{5pt}
}{}{}

\makeatother

\begin{document}

\title{Reliable Representations Make A Stronger Defender: Unsupervised Structure Refinement for Robust GNN}


\author{Kuan Li}
\authornotemark[2]
\affiliation{%
  \institution{Institute of Computing Technology, Chinese Academy of Sciences\\University of Chinese Academy of Sciences, Beijing, China}
  \city{}
  \country{}
}
\email{likuan_buaa@163.com}
\orcid{0000-0001-6303-7706}

\author{Yang Liu}
\authornotemark[2]
\affiliation{%
  \institution{Institute of Computing Technology, Chinese Academy of Sciences\\University of Chinese Academy of Sciences, Beijing, China}
  \city{}
  \country{}
}
\email{liuyang520ict@gmail.com}

\author{Xiang Ao}
\authornote{Corresponding author.}
\authornotemark[2]
\affiliation{%
  \institution{Institute of Computing Technology, Chinese Academy of Sciences\\ University of Chinese Academy of Sciences, Beijing, China}
  \city{}
  \country{}}
\email{aoxiang@ict.ac.cn}

\author{Jianfeng Chi\\ Jinghua Feng\\ Hao Yang}
\affiliation{%
  \institution{Alibaba Group, Hangzhou, China}
  \city{}
  \country{}
}
\email{jinghua.fengjh@alibaba-inc.com}
\email{youhiroshi.yangh@alibaba-inc.com}
\email{bianfu.cjf@alibaba-inc.com}

\author{Qing He}
\authornote{Key Lab of Intelligent Information Processing of Chinese Academy of Sciences (CAS). Xiang Ao is also at Institute of Intelligent Computing Technology, Suzhou, China.}
\affiliation{%
  \institution{Institute of Computing Technology, Chinese Academy of Sciences\\ University of Chinese Academy of Sciences, Beijing, China}
  \city{}
  \country{}
  \postcode{100190}
}
\email{heqing@ict.ac.cn}
\renewcommand{\shortauthors}{Kuan Li et al.}
\begin{abstract}
Benefiting from the message passing mechanism, Graph Neural Networks (GNNs) have been successful on flourish tasks over graph data. However, recent studies have shown that attackers can catastrophically degrade the performance of GNNs by maliciously modifying the graph structure. A straightforward solution to remedy this issue is to model the edge weights by learning a metric function between pairwise representations of two end nodes, which attempts to assign low weights to adversarial edges. The existing methods use either raw features or representations learned by supervised GNNs to model the edge weights. However, both strategies are faced with some immediate problems: raw features cannot represent various properties of nodes (\emph{e.g.}, structure information), and representations learned by supervised GNN may suffer from the poor performance of the classifier on the poisoned graph. We need representations that carry both feature information and as mush correct structure information as possible and are insensitive to structural perturbations. To this end, we propose an unsupervised pipeline, named STABLE, to optimize the graph structure. Finally, we input the well-refined graph into a downstream classifier. For this part, we design an advanced GCN that significantly enhances the robustness of vanilla GCN~\cite{kipf2017semi} without increasing the time complexity. Extensive experiments on four real-world graph benchmarks demonstrate that STABLE outperforms the state-of-the-art methods and successfully defends against various attacks. The implementation of STABLE is available at \bluefont{\href{https://github.com/likuanppd/STABLE}{https://github.com/likuanppd/STABLE}}

\end{abstract}



\maketitle          
\section{Introduction}
% \textls[-12]{Graphs are ubiquitous data structures that can represent various objects and their complex relations~\cite{zhang2020deep, battaglia2018relational, zhou2020graph}. As a powerful tool of learning representations from graphs, Graph Neural Networks~(GNNs) burgeon widely explorations in recent years~\cite{li2015gated, kipf2017semi, hamilton2017inductive, velivckovic2017graph, xu2019powerful}} for numerous graph-related tasks, primarily focused on node representation learning~\cite{perozzi2014deepwalk, kipf2016variational, velivckovic2018deep, you2020graph}, transductive node classifications~\cite{kipf2017semi, hamilton2017inductive, velivckovic2017graph}, and etc. 
% The key to the success of GNNs is the neural message passing mechanism, in which GNNs regard features and hidden representations as messages carried by nodes and propagate them through the edges in the graph. 

Graphs are ubiquitous data structures that can represent various objects and their complex relations~\cite{zhang2020deep, battaglia2018relational, zhou2020graph, liu2021pick, zhu2021intelligent}. As a powerful tool of learning representations from graphs, Graph Neural Networks~(GNNs) burgeon widely explorations in recent years~\cite{li2015gated, kipf2017semi, hamilton2017inductive, velivckovic2017graph, xu2019powerful} for numerous graph-based tasks, primarily focused on node representation learning~\cite{perozzi2014deepwalk, kipf2016variational, velivckovic2018deep, you2020graph} and transductive node classification~\cite{kipf2017semi, hamilton2017inductive, velivckovic2017graph, huang2022auc}. The key to the success of GNNs is the neural message passing mechanism, in which GNNs regard features and hidden representations as messages carried by nodes and propagate them through the edges. However, this mechanism also brings a security risk.

\begin{table}[t]
  \caption{The mean accuracy under different perturbation rates by MetaAttack on the Cora dataset. Here the perturbation rate is the ratio of changed edges. The data split follows 10\%/ 10\%/ 80/\%(train/ validation/ test).}
  \vspace{-2mm}
  \begin{tabular}{c|cccc}
    \toprule
    Ptb Rate&GCN&GRCN&GNNGuard&Jaccard\\
    \midrule
    0\% & 83.56 & \textbf{86.12} & 78.52 & 81.79 \\
    5\% & 76.36 &\textbf{80.78} & 77.96 & 80.23 \\
    10\% & 71.62 &72.42 & \textbf{74.86} & 74.65 \\
    20\% & 60.31 &65.43 & 72.03 & \textbf{73.11} \\
  \bottomrule
\end{tabular}
\label{motivation}
\end{table}

Recent studies have shown that GNNs are vulnerable to adversarial attacks~\cite{dai2018adversarial, zugner2018adversarial, zugner_adversarial_2019, wu2019adversarial, zhu2022binarizedattack}. In other words, by limitedly rewiring the graph structure or just perturbing a small part of the node features, attackers can easily fool the GNNs to misclassify nodes in the graph. The robustness of the model is essential for some security-critical domains. For instance, in fraudulent transaction detection, fraudsters can conceal themselves by deliberately dealing with common users. Thus, it is necessary to study the robustness of GNNs. Although attackers can modify the clean graph by perturbing node features or the graph structure, most of the existing adversarial attacks on graph data have concentrated on modifying graph structure~\cite{zugner_adversarial_2019, finkelshtein2020single}. Moreover, the structure perturbation is considered more effective~\cite{zugner2018adversarial, wu2019adversarial} probably due to the message passing mechanism. Misleading messages will pass through a newly added edge, or correct messages cannot be propagated because an original edge is deleted. In this paper, our purpose is to defend against the non-targeted adversarial attack on graph data that attempts to reduce the overall performance of the GNN. Under this setting, the GNNs are supposed to be trained on a graph that the structure has already been perturbed.

One representative perspective to defend against the attack is to refine the graph structure by reweighting the edges~\cite{zhu2021deep}. Specifically, edge weights are derived from learning a metric function between pairwise representations~\cite{zhang2020gnnguard, wang2020gcn, li2018adaptive, velivckovic2017graph, fatemi2021slaps}. Intuitively, the weight of an edge could be represented as a distance measure between two end nodes, and defenders can further prune or add edges according to such distances. 

%Specifically, edge weights are derived from learning a metric function between pairwise representations of two end nodes~\cite{zhang2020gnnguard, wang2020gcn, li2018adaptive, velivckovic2017graph, fatemi2021slaps}. Intuitively, the weight of an edge could be represented as a distance measure between two linked nodes, and defenders can further prune or add edges according to such distances. 

Though extensive approaches have been proposed to model the pairwise weights, most research efforts are devoted to designing a novel metric function, while the rationality of the inputs of the function is inadequately discussed. In more detail, they usually utilize the original features or representations learned by the supervised GNNs to compute the weights. 

% However, there are several drawbacks to using features or representations learned by the classifier. For example, GNNGuard~\cite{zhang2020gnnguard} and Jaccard\cite{wu2019adversarial} utilize cosine similarity of the initial features to model the edge weights, while GRCN~\cite{yu2020graph} uses the inner product of learned representations. The performance of these three models on Cora attacked by MetaAttack~\cite{zugner_adversarial_2019}\footnote{It is the state-of-the-art attack method that uses meta-gradients to maliciously modify the graph structure.} is listed in Table~\ref{motivation}. On the one hand, as the perturbation rate rises, the performance of GRCN drops rapidly. Since the representation is learned in a supervised fashion, poor classification performance will lead to poor representations, not optimising graph structure effectively. In other words, the quality of the representations co-vary with the task performance and is \textbf{sensitive to the perturbations}. On the other hand, the feature-based approaches, \emph{i.e.}, GNNGuard and Jaccard, appear insensitive to the perturbation rate changes, but apparently, there is a \textbf{trade-off between the performance and the robustness}. The vanilla GCN~\cite{kipf2017semi} even outperforms both GNNGuard and Jaccard in the clean graph. This is because the original features cannot sufficiently represent rich properties of nodes, such as structure information.

However, optimizing graph structures based on either features or supervised signals might not be reliable. For example, GNNGuard~\cite{zhang2020gnnguard} and Jaccard\cite{wu2019adversarial} utilize cosine similarity of the initial features to model the edge weights, while GRCN~\cite{yu2020graph} uses the inner product of learned representations. The performance of these three models on Cora attacked by MetaAttack~\cite{zugner_adversarial_2019}\footnote{It is the state-of-the-art attack method that uses meta-gradients to maliciously modify the graph structure.} is listed in Table~\ref{motivation}. 
From the table, we first observe feature-based methods do not perform well under low perturbation rates, because features cannot carry structural information. Optimizing the structure based on such an insufficient property can lead to mistakenly deleting normal edges~(the statistics of the removed edges is listed in Table \ref{remove}). 
When the perturbation is low, the negative impact of such mis-deletion is greater than the positive impact of deleting malicious edges.%, so the trade-off between robustness and correctness exists. 
Thus, we want to refine the structure by learned representations which contain structural information. 
Second, we also see the representations learned by the supervised GNNs are not reliable under high perturbations~(the results of GRCN). This is probably because attack methods are designed to degrade the accuracy of a surrogate GNN, so the quality of representations learned by the classifier co-vary with the task performance. 

Based on the above analysis, we consider that the representations used for structure refining should be obtained in a different manner, and two factors in terms of learning representations in the adversarial scenario should be highlighted: 1) \textbf{carrying feature information and in the meantime carrying as much correct structure information as possible} and 2) \textbf{insensitivity to structural perturbations}. 

%an unsupervised way. The main challenge from this perspective is how we can get reliable representations used for the edge pruning and adding. What we are concerned with are two factors in terms of learning representations in an adversarial scenario: 1) \textbf{stronger than the original features} and 2) \textbf{insensitive to structural perturbations}.  

% \textls[-20]{To tackle these issues, we propose an approach named \redfont{RGSR}~(\underline{R}obust \underline{G}NN} with Reliable \underline{S}tructure \underline{R}efinement) in this paper. First, RGSR prepossesses the graph to remove the easily detectable adversarial edges by calculating node features' similarities. Then we generate two augmentation views by randomly recovering a small portion of removed edges and devise a contrastive learning method to obtain node representations that are insensitive to intrinsic structural perturbations. Finally, such learned representation is utilized to perform graph structure refinement to derive an unpolluted graph. 
% Once the structure is well-refined, any GNNs can be used for the downstream learning tasks. By observing what makes edge editing a strong adversarial change, we find that GCN falls victim to its renormalization trick. Hence we also introduce an advanced message passing in the GCN module to further improve the robustness without hurting the efficiency. 
% Experiments on three widely used benchmarks demonstrate that our RGSR can clearly outperform the compared SOTA methods and the performance margin becomes significantly as the perturbation rate increases. 

To this end, we propose an approach named STABLE~(\underline{ST}ructure le\underline{A}rning GNN via more relia\underline{BL}e r\underline{E}presentations) in this paper, and it learns the representations used for structure refining by unsupervised learning. The unsupervised approach is relatively reliable because the objective is not directly attacked. Additionally, the unsupervised pipeline can be viewed as a kind of pretraining, and the learned representations may have been trained to be invariant to certain useful properties~\cite{wiles2021fine} (i.e., the perturbed structure here).
We design a contrastive method with a novel pre-process and recovery schema to obtain the representations. Different from the previous contrastive method~\cite{velivckovic2018deep, hassani2020contrastive, peng2020graph, Zhu:2020vf}, we roughly refine the graph to remove the easily detectable adversarial edges and generate augmentation views by randomly recovering a small portion of removed edges. Pre-processing makes the underlying structural information obtained during the 
representation learning process relatively correct, and such an augmentation strategy can be viewed as injecting slight attack to the pre-processed graph. Then the representations learned on different augmentation views tend to be similar during the contrastive training. That is to say, we obtain representations insensitive to the various slight attacks. Such learned representations fulfill our requirements and can be utilized to perform graph structure refinement to derive an unpolluted graph.

In addition, any GNNs can be used for the downstream learning tasks after the structure is well-refined. For this part, many methods~\cite{jin2020graph, jin2021node} just use the vanilla GCN~\cite{kipf2017semi}. By observing what makes edge insertion or deletion a strong adversarial change, we find that GCN falls victim to its renormalization trick. Hence we introduce an advanced message passing in the GCN module to further improve the robustness.

Our contributions can be summarized as follow:

% \begin{itemize}

%   \item We propose a contrastive method with robustness-oriented augmentations to obtain the representations used for structure refining, which can effectively capture structural information of nodes and are insensitive to the perturbations.
%   \item We further explore the reason for the lack of robustness of GCN and propose a more robust normalization trick.
%   \item Extensive experiments on four real-world datasets demonstrate that STABLE can defend against different types of adversarial attacks and outperform the state-of-the-art defense models.

% \end{itemize}  
(1) We propose a contrastive method with robustness-oriented augmentations to obtain the representations used for structure refining, which can effectively capture structural information of nodes and are insensitive to the perturbations. (2) We further explore the reason for the lack of robustness of GCN and propose a more robust normalization trick. (3) Extensive experiments on four real-world datasets demonstrate that STABLE can defend against different types of adversarial attacks and outperform the state-of-the-art defense models.

%  Even slight structural modifications can augment the representations. Hence, we consider an unsupervised approach. There are mainly three types of well-known graph-based unsupervised learning methods - random-walk~\cite{jeh2003scaling, perozzi2014deepwalk}, GAEs~\cite{kipf2016variational}, and contrastive learning~\cite{velivckovic2018deep, you2020graph}. GAEs and the random-walk are known to emphasize proximity at the expense of structural information~\cite{velivckovic2018deep, garcia2017learning}, and the augmentation scheme, like randomly edge perturbation, in contrastive methods are naturally similar to adversarial attacks~\cite{Zhu:2020vf, you2020graph}. Therefore, we introduce an unsupervised contrastive learning method with robustness-oriented structure augmentations to achieve the above requirements. Then based on the representations, we further optimize the structure.

\section{Preliminaries}
\subsection{Graph Neural Networks}
Let $\mathcal{G}=\{\mathcal{V}, \mathcal{E}, \mathbf{X}\}$ represent a graph with $N$ nodes, where $\mathcal{V}$ is the node set, $\mathcal{E}$ is the edge set, and $\mathbf{X} \in \mathbb{R}^{N\times d}$ is the original feature matrix of nodes. Let $\mathbf{A} \in\{0,1\}^{N\times N}$ represent the adjacency matrix of $\mathcal{G}$, in which $\mathbf{A}_{ij} \in\{0,1\}$ denotes the existence of the edge $e_{ij}\in\mathcal{E}$ that links node $v_i$ and $v_j$. 
The first-order neighborhood of node $v_i$ is denoted as $\mathcal{N}_i$, including node $v_i$ itself. We use $\mathcal{N}_i^*$ to indicate $v_i$'s neighborhood excluding itself. The transductive node classification task can be formulated as we now describe. Given a graph $\mathcal{G}$ and a subset $\mathcal{V}_L \subseteq \mathcal{V}$ of labeled nodes, with class labels from $\mathcal{C}=\{c_1, c_2,...,c_K\}$. $\mathcal{Y}_L$ and $\mathcal{Y}_U$ denote the ground-truth labels of labeled nodes and unlabeled nodes, respectively. The goal is to train a GNN $f_\theta$ to learn a function: $\mathcal{V} \rightarrow \mathcal{Y}$ that maps the nodes to the label set so that $f_\theta$ can predict labels of unlabeled nodes. $\theta$ is the trainable parameters of GNN. 

GNNs can be generally specified as $f_\theta(\mathbf{X}, \mathbf{A})$~\cite{zhou2020graph, kipf2017semi, hamilton2017inductive}. Each layer can be divided into a message passing function ($\textrm{MSP}$) and an updating function ($\textrm{UPD}$). Given a node $v_i$ and its neighborhood $\mathcal{N}_i^*$, GNN first implements $\textrm{MSP}$ to aggregate information from $\mathcal{N}_i^*$: $\bm{m}_i^t = \textrm{MSP}(\{\bm{h}_j^{t-1};j\in \mathcal{N}_i^*\})$, where $\bm{h}_j^{t-1}$ denotes the hidden representation in the previous layer, and $\bm{h}_j^{0}=\bm{x}_j$. Then, GNN updates the representation by $\textrm{UPD}$: $\bm{h}_i^t = \textrm{UPD}(\bm{m}_i^t, \bm{h}_i^{t-1})$, which is usually a sum or concat function. GNNs can be designed in an end-to-end fashion and can also serve as a representation learner for downstream tasks~\cite{kipf2017semi}.

\subsection{Gray-box Poisoning Attacks and Defence}
In this paper, we explore the robustness of GNNs under non-targeted Gray-box poisoning attack. Gray-box~\cite{yu2020graph} means the attacker holds the same data information as the defender, and the defense model is unknown. Poisoning attack represents GNNs are trained on a graph that attackers maliciously modify, and the aim of it is to find an optimal perturbed $\mathbf{A}'$, which can be formulated as a bilevel optimization problem~\cite{zugner2018adversarial,zugner_adversarial_2019}:
\begin{equation}
\begin{split}
  \underset{A'\in \Phi (A)}{\textrm{argmin}} \quad \mathcal{L}_{atk}(f_{\theta^*}(\mathbf{A}', \mathbf{X})) \quad\\
  s.t. \quad \theta^*=\underset{\theta}{\textrm{argmin}}\ \mathcal{L}_{train}(f_{\theta}(\mathbf{A}', \mathbf{X})).
\end{split}
\end{equation}
Here $\Phi (\mathbf{A})$ is a set of adjacency matrix that fit the constraint: $\frac{\left\|\mathbf{A}' - \mathbf{A} \right \|_0}{\left\|\mathbf{A}\right\|_0} \leq \Delta$, $\mathcal{L}_{atk}$ is the attack loss function, and $\mathcal{L}_{train}$ is the training loss of GNN. The $\theta^*$ is the optimal parameter for $f_\theta$ on the perturbed graph. $\Delta$ is the maximum perturbation rate. In the non-targeted attack setting, the attacker aims to degrade the overall performance of the classifier. 

Many efforts have been made to improve the robustness of GNNs ~\cite{xu2019topology, zhu2019robust, entezari2020all, jin2020graph, liu2021elastic, jin2021node}. Among them, metric learning approach~\cite{zhu2021deep} is a promising method~\cite{li2018adaptive, zhang2020gnnguard, wang2020gcn, chen2020iterative}, which refines the graph structure by learning a pairwise metric function $\phi(\cdot, \cdot)$:
\begin{equation}
    \textbf{S}_{ij}=\phi(\bm{z}_i, \bm{z}_j),
\end{equation}
where $\bm{z}_i$ is the raw feature or the learned embedding of node $v_i$ produced by GNN, and $\textbf{S}_{ij}$ denotes the learned edge weight between node $v_i$ and $v_j$. $\phi(\cdot, \cdot)$ can be a similarity metric or implemented in multilayer perceptions (MLPs). Further, the structure can be optimized according to the weights matrix $\mathbf{S}$. For example, GNNGuard~\cite{zhang2020gnnguard} utilizes cosine similarity to model the edge weights and then calculates edge pruning probability through a non-linear transformation. Moreover, GRCN~\cite{yu2020graph} and GAUGM~\cite{zhao2021data} directly compute the weights of the edges by the inner product of embeddings, and no additional parameters are needed:
\begin{equation}
  \textbf{S}_{ij}=\sigma(\mathbf{ZZ}^{\mathrm{T}}).
\end{equation}
Such methods aim to assign high weights to helpful edges and low weights to adversarial edges or even remove them. Here we want to define what is a helpful edge. Existing literature posits that strong homophily of the underlying graph is necessary for GNNs to achieve good performance on transductive node classification~\cite{abu2019mixhop, chien2020adaptive, hou2019measuring}. Under the homophily assumption~\cite{mcpherson2001birds}, the edge which links two similar nodes (\emph{e.g.}, same label) might help the nodes to be correctly classified. However, since only few labeled nodes are available, previous works utilize the raw features or the representations learned by the task-relevant GNNs to search for similar nodes.

Different from the aforementioned defense methods, we leverage an unsupervised pipeline to calculate the weights matrix and refine the graph structure.

\begin{figure*}[h]
  \includegraphics[width=\textwidth]{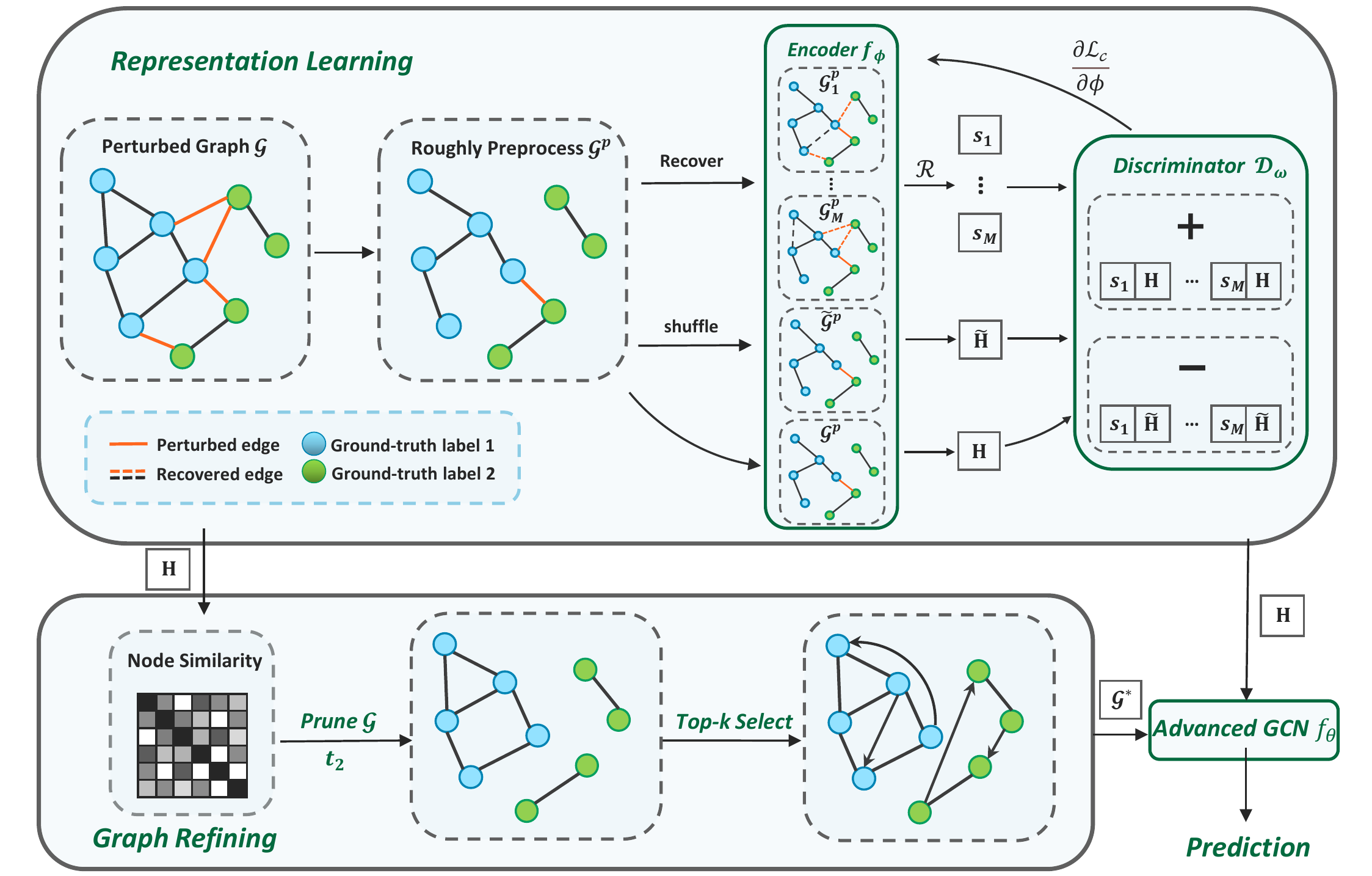}
  \vspace{-6mm} 
  \caption{Overall framework. Dash lines mean recovered edges.}
  \vspace{-2mm}
  % This figure demonstrates the overall framework of STABLE. Input is a perturbated graph. First, we pre-process the graph by removing some edegs that might be perturbations. The pre-processed graph is denoted as $\mathcal{G}^P$. Then we generate two robustness-oriented augmentation views, $\mathcal{G}^P_1$ and $\mathcal{G}^P_2$, by randomly recovering a small portion of the removed edegs. Further a feature-shuffled graph $\widetilde{\mathcal{G}}^P$is used to sample negative paris in contrastive learning. We train an encoder and a discriminator to maximize the mutual information between the local representation in $\mathcal{G}^P$ and the global representation in $\mathcal{G}_1^P$ and $\mathcal{G}_2^P$. The representation $\textbf{H}$ learned by the encoder on $\mathcal{G}^P$ will be used to refine the graph structure. Finally, we leverage an advanced GCN for the downstream task, which is transductive node classffication in this work.
  \label{model}
\end{figure*}

\section{Method}
In this section, we will present STABLE in a top-down fashion: starting with an elaboration of the overall framework, followed by the specific details of the structural optimization module, and concluding by an exposition of the advanced GCN.

\subsection{The Overall Architecture}
The illustration of the framework is shown in Fig. \ref{model}. It consists of two major components, the structure learning network and the advanced GCN classifier. The structure learning network can be further divided into two parts, representation learning and graph refining. For the representation learning part, we design a novel contrastive learning method with a robustness-oriented graph data augmentation, which randomly recovers the roughly removed edges. Then the learned representations are utilized to revise the structure according to the homophily assumption~\cite{mcpherson2001birds}. Finally, in the classification step, by observing what properties the nodes connected by the adversarial edges have, we carefully change the renormalization trick in GCN\cite{kipf2017semi} to improve the robustness. The overall training algorithm is shown in Appendix \ref{algorithm}. 

\subsection{Representation learning}
\label{representationlearning}
According to \cite{zugner_adversarial_2019, wu2019adversarial}, most of the perturbations are edge insertions, which connect nodes with different labels. Thus a straightforward method to deal with the structural perturbation is to find the adversarial edges and remove them. Previous works utilize the raw features or the representations learned by the end-to-end GNNs to seek the potential perturbations. As we mentioned before, such approaches have flaws.

% \begin{table}[h]
%   \caption{Frequency of Special Characters}
%   \label{motivation}
%   \begin{tabular}{c|cccc}
%     \toprule
%     $\Delta$&0\%&5\%&10\%&20\%\\
%     \midrule
%     GRCN & \textbf{86.12} & \textbf{80.78} & 72.42 &65.43\\
%     GNNGuard & 78.52 & 77.96 & \textbf{74.86} & 72.03\\
%     Jaccard & 81.79 & 80.23 & 74.65 &\textbf{73.11}\\
%   \bottomrule
% \end{tabular}
% \end{table}

STABLE avoids the pitfalls by leveraging a task-irrelevant contrastive method with robustness-oriented augmentations to learn the representations. 
% The unsupervised methods on the graph can be roughly divided into three categories: GAEs, random walk methods, and contrastive methods, and both GAEs and random walk methods are suffered from the proximity over-emphasizing\cite{velivckovic2018deep}. Inspired by the success of graph contrastive methods\cite{velivckovic2018deep, you2020graph, Zhu:2020vf, hassani2020contrastive}, we design a robustness-oriented augmentation method for adversarial attack scenarios.
First, we roughly pre-process the perturbed graph based on a similarity measure, \emph{e.g.}, Jaccard similarity or cosine similarity. We quantify the score $\textbf{S}_{ij}$ between node $v_i$ and its neighbor $v_j$ as follows:
\begin{equation}
  \textbf{S}_{ij}=\textrm{sim}(\bm{x}_i, \bm{x}_j),
  \label{pre-process}
\end{equation}
where $\mathcal{S}$ is the score matrix, $\bm{x}_i$ and $\bm{x}_j$ are the feature vectors of $v_i$ and $v_j$, respectively. Then we prune the edges whose scores are below a threshold $t_1$. The removed edge matrix and roughly pre-processed graph are denoted as $\textbf{E}$ and $\mathcal{G}^P$, where $\textbf{E}_{ij}=1$ denotes the edge between $v_i$ and $v_j$ is removed.

Generating views is the key component of contrastive methods. Different views of a graph provide different contexts for each node, and we hope our views carry the context for enhancing the robustness of the representations. Previous works generate augmentations by randomly perturbing the structure~\cite{Zhu:2020vf, you2020graph} on the initial graph. Different from them, we generate $M$ graph views, denoted as $\mathcal{G}_1^P$ to $\mathcal{G}_M^P$, by randomly recovering a small portion of $\textbf{E}$ on the pre-processed graph $\mathcal{G}^P$. Formally, we first sample a random masking matrix $\textbf{P} \in \{0, 1\}^{N\times N}$, whose entry is drawn from a Bernoulli distribution $\textbf{P}_{ij} \sim \mathcal{B}(1 - p)$ if $\textbf{E}_{ij}=1$ and $\textbf{P}_{ij}=0$ otherwise. Here $p$ is a hyper-parameter that controls the recovery ratio. The adjacency matrix of the augmentation view can be computed as: 
\begin{equation}
  \mathbf{A}_1^P= \mathbf{A}^P + \textbf{E} \odot \textbf{P},
  \label{augmentation}
\end{equation}
where $\mathbf{A}^P$ and $\mathbf{A}_1^P$ denote the adjacency matrix of $\mathcal{G}^P$ and $\mathcal{G}_1^P$ respectively, and $\odot$ is the Hadamard product. Other views can be obtained in the same way. The pre-process and the robustness-oriented augmentations are critical to the robustness, and we will elaborate on this point at the end of this subsection.

Our objective is to learn a one-layer GCN encoder $f_\phi$ parameterized by $\phi$ to maximize the mutual information between the local representation in $\mathcal{G}^P$ and the global representations in $\mathcal{G}_j$. Here $\mathcal{G}_j$ denote arbitrary augmentation $j$. The encoder follows the propagation rule:
\begin{equation}
  f_\phi(\mathbf{X},\mathbf{A})=\sigma(\hat{\mathbf{A}}\mathbf{X}\textbf{W}_\phi)
  \label{encoder}
\end{equation}
where $\hat{\mathbf{A}}=(\textbf{D}+\textbf{I}_N)^{-\frac{1}{2}}(\mathbf{A}+\textbf{I}_N)(\textbf{D}+\textbf{I}_N)^{-\frac{1}{2}}$ is the renormalized adjacency matrix.
we denote node representations in $\mathcal{G}^P$ and augmentation $j$ as $\textbf{H}=f_\phi (\mathbf{X}, \mathbf{A}^P)$ and $\textbf{H}_j=f_\phi (\mathbf{X}, \mathbf{A}_j^P)$. The global representation can be obtained via a readout function $\mathcal{R}$. We leverage a global average pooling function:
\begin{equation}
  \bm{s}_j = \sigma \left( \frac{1}{N}\sum_{i=1}^{N}\bm{h}_{i,j} \right),
  \label{readout}
\end{equation}
where $\sigma$ is the logistic sigmoid nonlinearity, and $\bm{h}_{i, j}$ is the representation of node $v_i$ in $\mathcal{G}_j^P$. We employ a discriminator $\mathcal{D}_\omega$ to distinguish positive samples and negative samples, and the output $\mathcal{D}_\omega(\bm{h}, \bm{s})$ represents the probability score of the sample pair from the joint distribution. $\omega$ is the parameters of the discriminator. To generate negative samples, which means 
the pair sampled from the product of marginals, we randomly shuffle the nodes' features in $\mathcal{G}^P$. We denote the shuffled graph and its representation as $\widetilde{\mathcal{G}}^P$ and $\widetilde{\textbf{H}}=f_\phi (\widetilde{\mathbf{X}}, \mathbf{A}^P)$. To maximize the mutual information, we use a binary cross-entropy loss between the positive samples and negative samples, and the objective can be writen as:
\begin{equation}
  \begin{split}
  \mathcal{L}_C=-\frac{1}{2N}\sum_{i=1}^N\Biggl(\frac{1}{M}\sum_{j=1}^M\biggl(\textrm{log}\mathcal{D}_\omega(\bm{h}_i, \bm{s}_j) + \textrm{log}\Bigl(1 - \mathcal{D}_\omega(\tilde{\bm{h}}_i, \bm{s}_j)\Bigr)\biggr) \Biggr).
  \end{split}
\label{lossc}
\end{equation}
where $\bm{h}_i$ and $\tilde{\bm{h}}_i$ are the representations of node $v_i$ in $\mathcal{G}^P$ and $\widetilde{\mathcal{G}}^P$. Minimizing this objective function can effectively maximize the mutual information between the local representations in $\mathcal{G}^P$ and the global representations in the augmentations based on the Jensen-Shannon MI (mutual information) estimator\cite{hjelm2018learning}.

To clarify why pre-processing and the designed augmentation method are crucial to robustness, we count the results of the pre-process and the recovery: Using Jaccard as the measure of the rough pre-process, $1,705$ edges are removed on Cora under 20\% perturbation rate. Among them, 691 are adversarial edges, and 1014 are normal edges. 259 edges out of $1,014$ connect two nodes with different labels. More than half of the removals are helpful. Therefore, most of the edges in $\textbf{E}$ can be regarded as adversarial edges so that \textbf{the recovery can be viewed as injecting slight attacks on $\mathcal{G}^P$}. The degrees of perturbation can be ranked as $\mathcal{G}\gg\mathcal{G}_1^P\approx\mathcal{G}_2^P\ ... \approx\mathcal{G}_M^P>\mathcal{G}^P$.

Recall that we need representations that carry as much correct structural information as possible and are insensitive to perturbations. We train our encoder on the much cleaner graphs, \emph{i.e.}, $\mathcal{G}^P$ and the views,  so the representations are much better than those learned in the original graph. The process of maximizing mutual information makes the representations learned in $\mathcal{G}^P$ and in the views similar so that the representations will be insensitive to the recovery. In other words, they will be insensitive to perturbations. In short, the rough pre-process meets the former requirement, and the robustness-oriented augmentation meets the latter.

\subsection{Graph Refining}
Once the high-quality representations are prepared, we can further refine the graph structure in this section. Existing literature posits that strong homophily of the underlying graph can help GNNs to achieve good performance on transductive node classification~\cite{abu2019mixhop, chien2020adaptive, hou2019measuring}. Therefore, the goal of the refining is to reduce heterophilic edges~(connecting two nodes in the different classes) and add homophilic edges~(connecting two nodes in the same class). Under homophily assumption~\cite{mcpherson2001birds}, similar nodes are more likely in the same class, so we use node similarity to search for potential homophilic edges and heterophilic edges.

We leverage the representations learned by the contrastive learning encoder to measure the nodes similarity:
\begin{equation}
  \textbf{M}_{ij}=\textrm{sim}(\bm{h}_i, \bm{h}_j),
  \label{similarity}
\end{equation}
where $\textbf{M}$ is the similarity score matrix, and $\textbf{M}_{ij}$ is the score of node $v_i$ and node $v_j$. Here we utilize the cosine similarity. Then we remove all the edges that connect nodes with similarity scores below a threshold $t_2$. The above process can be formulated as:
\begin{equation}
  \mathbf{A}^R_{ij}=\left\{
  \begin{aligned}
    1 \quad& if\ \textbf{M}_{ij} > t_2\ and\ \mathbf{A}_{ij}=1 \\
    0 \quad& otherwise,
  \end{aligned}
  \right.
  \label{prune}
  \end{equation}
where $\mathbf{A}^R$ is the adjacency matrix after pruning.

After pruning, we insert some helpful edges by building a top-$k$ matrix $\textbf{T}^k$. For each node, we connect it to $k$ nodes that are most similar to it. Formally, $\textbf{T}^k_{ij}=1$, if $v_j$ is one of the $k$ nodes most similar to $v_i$. Note that $\textbf{T}^k$ is not a symmetric matrix because the similarity relation is not symmetric. For example, $v_i$ is the most similar node to $v_j$ but not vise versa. The pruning strategy cannot eliminate all the perturbations, and these insertions can effectively diminish the impact of the remaining harmful edges. From our empirical results, this is particularly effective when the perturbation rate is high. The optimal adjacency matrix can be formulated as:
\begin{equation}
  \mathbf{A}^*=\mathbf{A}^R+\textbf{T}^k
  \label{addedge}
\end{equation}
The optimal graph is denoted as $\mathcal{G}^*$, and it is a directed graph due to the asymmetry of $\textbf{T}$. 

\subsection{Advanced GCN In the Adversarial Attack Scenario}
After the graph is well-refined, any GNNs can be used for the downstream task. We find that only one modification to the renormalization trick in GCN is needed to enhance the robustness greatly. This improvement does not introduce additional learnable parameters and therefore does not reduce the efficiency of the GCN.

\begin{figure}[h]
%   \vspace{-0.4cm}
%   \setlength{\abovecaptionskip}{0.1cm} 
%   \setlength{\belowcaptionskip}{-0.4cm}
  \centering
  \includegraphics[width=\linewidth]{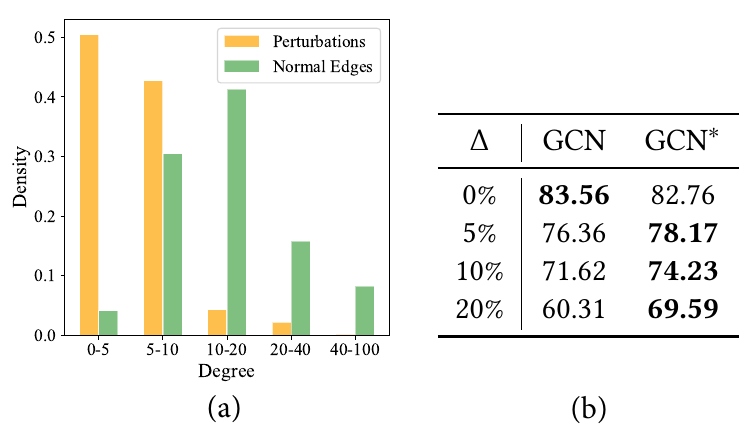}
  \vspace{-7mm}
  \caption{(a): The edge degree distribution of the clean graph and the distribution of the adversarial edges on Cora attacked by MetaAttack. (b): The mean accuracy under different perturbation rates on Cora. The data split follows 10\%/ 10\%/ 80/\%(train/ val/test).} 
  \Description{A toy example}
  \label{advanced}
  \vspace{-3mm}
\end{figure}

\cite{zugner_adversarial_2019} studied the node degree distributions of the clean graph and the node degrees of the nodes that are picked for adversarial edges. Not surprisingly, the nodes selected by the adversarial edges are those with low degrees. The nodes with only a few neighbors are more vulnerable than high-degree nodes due to the message passing mechanism. However, the discussion about the picked nodes only exposes one-side property of the adversarial edges. We further define the degree of an edge  as the sum of both connected nodes' degrees. The edge degree distribution is shown in Fig. \ref{advanced} (a). It shows that adversarial edges tend to link \textbf{two} low-degree nodes, and nearly half of them are below 5 degrees.
% \begin{figure}[t]
%   \begin{minipage}{0.48\linewidth}
%    \centerline{\includegraphics[width=4cm]{degree.pdf}}
%    \label{distribution}
%   \end{minipage}
%   \hfill
%   \begin{minipage}{0.48\linewidth}
%     \begin{tabular}{c|cc}
%       \toprule
%       $\Delta$&GCN&$\textrm{GCN}^*$\\
%       \midrule
%       0\% & \textbf{83.56} & 82.76\\
%       5\% & 76.36 &\textbf{78.17}\\
%       10\% & 71.62 &\textbf{74.23}\\
%       20\% & 60.31 &\textbf{69.59}\\
%     \bottomrule
%   \end{tabular}
%   \end{minipage}
%   \caption{Left: The edge degree distribution of the clean graph and the distribution of the adversarial edges on Cora attacked by MetaAttack. Right: The mean accuracy under different perturbation rate on Cora. The data split follows 10\%/ 10\%/ 80/\%(train/ val/test).}
%   \label{advanced}
% \end{figure}
 
GCN assigns higher aggregation weights to low-degree neighbors, and the renormalization adjacency matrix is formulated as: $\hat{\mathbf{A}}=(\textbf{D}+\textbf{I}_N)^{-\frac{1}{2}}(\mathbf{A}+\textbf{I}_N)(\textbf{D}+\textbf{I}_N)^{-\frac{1}{2}}$. However, based on the above analysis, low-degree neighbors are more likely to be maliciously added neighbors. To give a specific example, here is node $v_i$ with only one neighbor in the clean graph, and the attacker link it to a low-degree node. In the aggregation step, only half of the messages are are correct. Even worse, the misleading message is assigned a higher weight because the added neighbor has a lower degree than the original neighbor.

Hence, a simple but effective way to improve the robustness of GCN is to assign lower weights to the low-degree neighbors. We propose our advanced GCN, which will implement the message passing as follow:
\begin{equation}
  \bm{h}_i^t = \textrm{ReLU}\Bigl( \bigl(\sum_{j \in \mathcal{N}_i^*} \frac{(d_id_j)^\alpha}{Z}\bm{h}_j^{t-1} + \beta \bm{h}_i^{(t-1)}\bigr)\mathbf{W}_\theta^t\Bigr)
  \label{prop}
\end{equation}
where $\alpha$ is the hyper-parameter that controls the weight design scheme, $\beta$ is the hyper-parameter that controls the weight of self-loop, $Z$ is a normalization coefficient, and $\textrm{ReLU}(\cdot)=max(0, \cdot)$. $h_i^0$ is the representation learned in Section \ref{representationlearning}. When $\alpha$ is large, a node aggregate more from the high-degree nodes. $\beta$ is also designed for robustness. Once the neighbors are unreliable, nodes should refer more to themselves.

High-degree nodes can be viewed as robust nodes, because a few perturbations can hardly cancel out the helpful information brought by their original neighbors. The learnable attack algorithms tend to bypass them. Thus, the message from these nodes are usually more reliable.

We show an example in Fig. \ref{advanced} (b). $\textrm{GCN}^*$ denotes the advanced GCN with $\alpha = 0.6$ and $\beta = 2$. As the perturbation rate increases, the robustness improves more obviously. In fact, this normalization trick can merge with any GNNs.

We train an advanced GCN $f_\theta$ by minimizing the cross-entropy:
\begin{equation}
  \mathcal{L}_G= \sum_{i \in \mathcal{V}_L}\ell(f_\theta(\textbf{H}, \mathbf{A}^*)_i, y_i)
  \label{lossg}
\end{equation}
where $\ell$ is the cross entropy loss. 

\begin{table*}[t]
%   \vspace{0.1cm}
%   \setlength{\abovecaptionskip}{0.1cm}
%   \setlength{\belowcaptionskip}{-0.4cm}
  \caption{Classification accuracy(\%) under different perturbation rates. The top two performance is highlighted in bold and underline.}
  \vspace{-2mm}
  \setlength{\tabcolsep}{1mm}{
  \begin{tabular}{c|c|ccccccccc}
    \toprule
    Dataset & Ptb Rate & GCN & RGCN & Jaccard & GNNGuard & GRCN & ProGNN & SimPGCN & Elastic & STABLE\\
    \midrule
    \multirow{5}*{Cora} & 0\% & 83.56$\pm$0.25 & 83.85$\pm$0.32 & 81.79$\pm$0.37 & 78.52$\pm$0.46 & \textbf{86.12$\pm$0.41} & 84.55$\pm$0.30 & 83.77$\pm$0.57 & 84.76$\pm$0.53 & \underline{85.58$\pm$0.56} \\
                        & 5\% & 76.36$\pm$0.84 & 76.54$\pm$0.49 & 80.23$\pm$0.74 & 77.96$\pm$0.54 & 80.78$\pm$0.94 & 79.84$\pm$0.49 & 78.98$\pm$1.10 & \textbf{82.00$\pm$0.39} & \underline{81.40$\pm$0.54} \\
                        & 10\% & 71.62$\pm$1.22 & 72.11$\pm$0.99 & 74.65$\pm$1.48 & 74.86$\pm$0.54 & 72.43$\pm$0.78 & 74.22$\pm$0.31 & 75.07$\pm$2.09 & \underline{76.18$\pm$0.46} & \textbf{80.49$\pm$0.61} \\
                        & 15\% & 66.37$\pm$1.97 & 65.52$\pm$1.12 & 74.29$\pm$1.11 & 74.15$\pm$1.64 & 70.72$\pm$1.13 & 72.75$\pm$0.74 & 71.42$\pm$3.29 & \underline{74.41$\pm$0.97} & \textbf{78.55$\pm$0.44} \\
                        & 20\% & 60.31$\pm$1.98 & 63.23$\pm$0.93 & \underline{73.11$\pm$0.88} & 72.03$\pm$1.11 & 65.34$\pm$1.24 & 64.40$\pm$0.59 & 68.90$\pm$3.22 & 69.64$\pm$0.62 & \textbf{77.80$\pm$1.10} \\
    \midrule
    \multirow{5}*{Citeseer} & 0\% & 74.63$\pm$0.66 & 75.41$\pm$0.20 & 73.64$\pm$0.35 & 70.07$\pm$1.31 & \underline{75.65$\pm$0.21} & 74.73$\pm$0.31 & 74.66$\pm$0.79 & 74.86$\pm$0.53 & \textbf{75.82$\pm$0.41} \\
                        & 5\% & 71.13$\pm$0.55 & 72.33$\pm$0.47 & 71.15$\pm$0.83 & 69.43$\pm$1.46 & \textbf{74.47$\pm$0.38} & 72.88$\pm$0.32 & 73.54$\pm$0.92 & 73.28$\pm$0.59 & \underline{74.08$\pm$0.58} \\
                        & 10\% & 67.49$\pm$0.84 & 69.80$\pm$0.54 & 69.85$\pm$0.77 & 67.89$\pm$1.09 & 72.27$\pm$0.69 & 69.94$\pm$0.45 & 72.03$\pm$1.30 & \underline{73.41$\pm$0.36} & \textbf{73.45$\pm$0.40} \\
                        & 15\% & 61.59$\pm$1.46 & 62.58$\pm$0.69 & 67.50$\pm$0.78 & 69.14$\pm$0.84 & 67.48$\pm$0.42 & 62.61$\pm$0.64 & \underline{69.82$\pm$1.67} & 67.51$\pm$0.45 & \textbf{73.15$\pm$0.53} \\
                        & 20\% & 56.26$\pm$0.99 & 57.74$\pm$0.79 & 67.01$\pm$1.10 & 69.20$\pm$0.78 & 63.73$\pm$0.82 & 55.49$\pm$1.50 & \underline{69.59$\pm$3.49} & 65.65$\pm$1.95 & \textbf{72.76$\pm$0.53} \\
    \midrule
    \multirow{5}*{Polblogs} & 0\% & 95.04$\pm$0.11 & 95.38$\pm$0.14 & / & / & 94.89$\pm$0.24 & \underline{95.93$\pm$0.17} & 94.86$\pm$0.46 & 95.57$\pm$0.26 & \textbf{95.95$\pm$0.27} \\
                        & 5\% & 77.55$\pm$0.77 & 76.46$\pm$0.47 & / & / & 80.37$\pm$0.46 & \underline{93.48$\pm$0.54} & 75.08$\pm$1.08 & 90.08$\pm$1.06 & \textbf{93.80$\pm$0.12} \\
                        & 10\% & 70.40$\pm$1.13 & 70.35$\pm$0.40 & / & / & 69.72$\pm$1.36 & \underline{85.81$\pm$1.00} & 68.36$\pm$1.88 & 84.05$\pm$1.94 & \textbf{92.46$\pm$0.77} \\
                        & 15\% & 68.49$\pm$0.49 & 67.74$\pm$0.50 & / & / & 66.56$\pm$0.93 & \underline{75.60$\pm$0.70} & 65.02$\pm$0.74 & 72.17$\pm$0.74 & \textbf{90.04$\pm$0.72} \\
                        & 20\% & 68.47$\pm$0.54 & 67.31$\pm$0.24 & / & / & 68.20$\pm$0.71 & \underline{73.66$\pm$0.64} & 64.78$\pm$1.33 & 71.76$\pm$0.92 & \textbf{88.46$\pm$0.33} \\
    \midrule
    \multirow{5}*{Pubmed} & 0\% & 86.83$\pm$0.06 & 86.02$\pm$0.08 & 86.85$\pm$0.09 & 85.24$\pm$0.07 & 86.72$\pm$0.03 & 87.33$\pm$0.18 & \textbf{88.12$\pm$0.17} & 87.71$\pm$0.06 & \underline{87.73$\pm$ 0.11} \\
                        & 5\% & 83.18$\pm$0.06 & 82.37$\pm$0.12 & 86.22$\pm$0.08 & 84.65$\pm$0.09 & 84.85$\pm$0.07 & \underline{87.25$\pm$0.09} & 86.96$\pm$0.18 & 86.82$\pm$0.13 & \textbf{87.59$\pm$0.08} \\
                        & 10\% & 81.24$\pm$0.17 & 80.12$\pm$0.12 & 85.64$\pm$0.08 & 84.51$\pm$0.06 & 81.77$\pm$0.13 & \underline{87.25$\pm$0.09} & 86.41$\pm$0.34 & 86.78$\pm$0.11 & \textbf{87.46$\pm$0.12} \\
                        & 15\% & 78.63$\pm$0.10 & 77.33$\pm$0.16 & 84.57$\pm$0.11 & 84.78$\pm$0.10 & 77.32$\pm$0.13 & \underline{87.20$\pm$0.09} & 85.98$\pm$0.30 & 86.36$\pm$0.14 & \textbf{87.38$\pm$0.09} \\
                        & 20\% & 77.08$\pm$0.2 & 74.96$\pm$0.23 & 83.67$\pm$0.08 & 84.25$\pm$0.07 & 69.89$\pm$0.21 & \underline{87.09$\pm$0.10} & 85.62$\pm$0.40 & 86.04$\pm$0.17 & \textbf{87.24$\pm$0.08} \\

    \bottomrule
  \end{tabular}}
  \label{main}
\end{table*}

\section{Experiments}
In this section, we empirically analyze the robustness and effectiveness of STABLE against different types of adversarial attacks. Specifically, we aim to answer the following research questions:
\begin{itemize}
  \item $\mathbf{RQ1}$: Does STABLE outperform the state-of-the-art defense models under different types of adversarial attacks?
  \item $\mathbf{RQ2}$: Is the structure learned by STABLE better than learned by other methods?
  \item $\mathbf{RQ3}$: What is the performance with respect to different training parameters?
  \item $\mathbf{RQ4}$: How do the key components benefit the robustness? (see Appendix \ref{secablation})
\end{itemize}

\subsection{Experimental Settings}

\subsubsection{Datasets} Following ~\cite{zugner2018adversarial, jin2020graph, entezari2020all}, we conduct our experiments on four benchmark datasets, including three citation graphs, \emph{i.e.}, Cora, Citeseer, and PubMed, and one blog graph, \emph{i.e.}, Polblogs. The statistics of these datasets are shown in Appendix \ref{dataset}.

\subsubsection{Baselines}
%提一句为什么没和GIB、NRGNN之类的比，因为他们防御的是其他类型的攻击，比如targeted-attack，label noise。
We compare STABLE with representative and state-of-the-art defence GNNs to verify the robustness. The baselines and SOTA approaches are \textbf{GCN}~\cite{kipf2017semi}, \textbf{RGCN}~\cite{zhu2019robust}, \textbf{Jaccard}~\cite{wu2019adversarial}, \textbf{GNNGuard}~\cite{zhang2020gnnguard}, \textbf{GRCN}~\cite{yu2020graph}, \textbf{ProGNN}~\cite{jin2020graph}, \textbf{SimpGCN}~\cite{jin2021node}, \textbf{Elastic}~\cite{liu2021elastic}. 
We implement three non-targeted structural adversarial attack methods, \emph{i.e.}, \textbf{MetaAttack~\cite{zugner_adversarial_2019}}, \textbf{DICE~\cite{waniek2018hiding}} and \textbf{Random}.

We introduce all these defence methods and attack methods in the Appendix \ref{baselines intro}.

\subsubsection{Implementation Details} 
The implementation details and parameter settings are introduced in Appendix \ref{implementation}
\begin{figure*}[t]
  \centering
  \subfigure[$\textrm{Cora}_{\textrm{DICE}}$]{
  \begin{minipage}[t]{0.24\textwidth}
  \centering
  \includegraphics[width=1.7in]{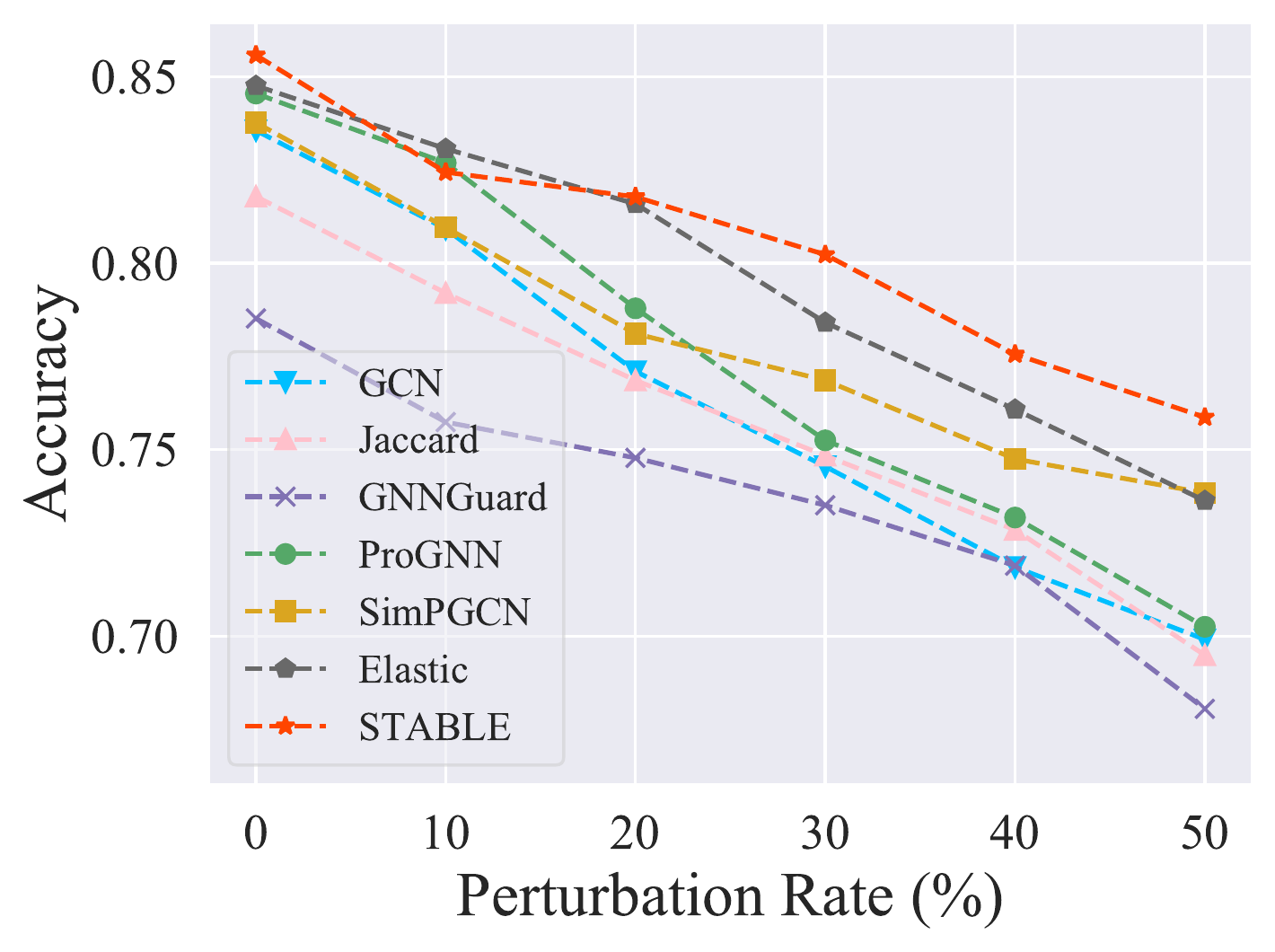}
  %\caption{fig1}
  \end{minipage}%
  }%
  \subfigure[$\textrm{Citeseer}_{\textrm{DICE}}$]{
  \begin{minipage}[t]{0.24\textwidth}
  \centering
  \includegraphics[width=1.7in]{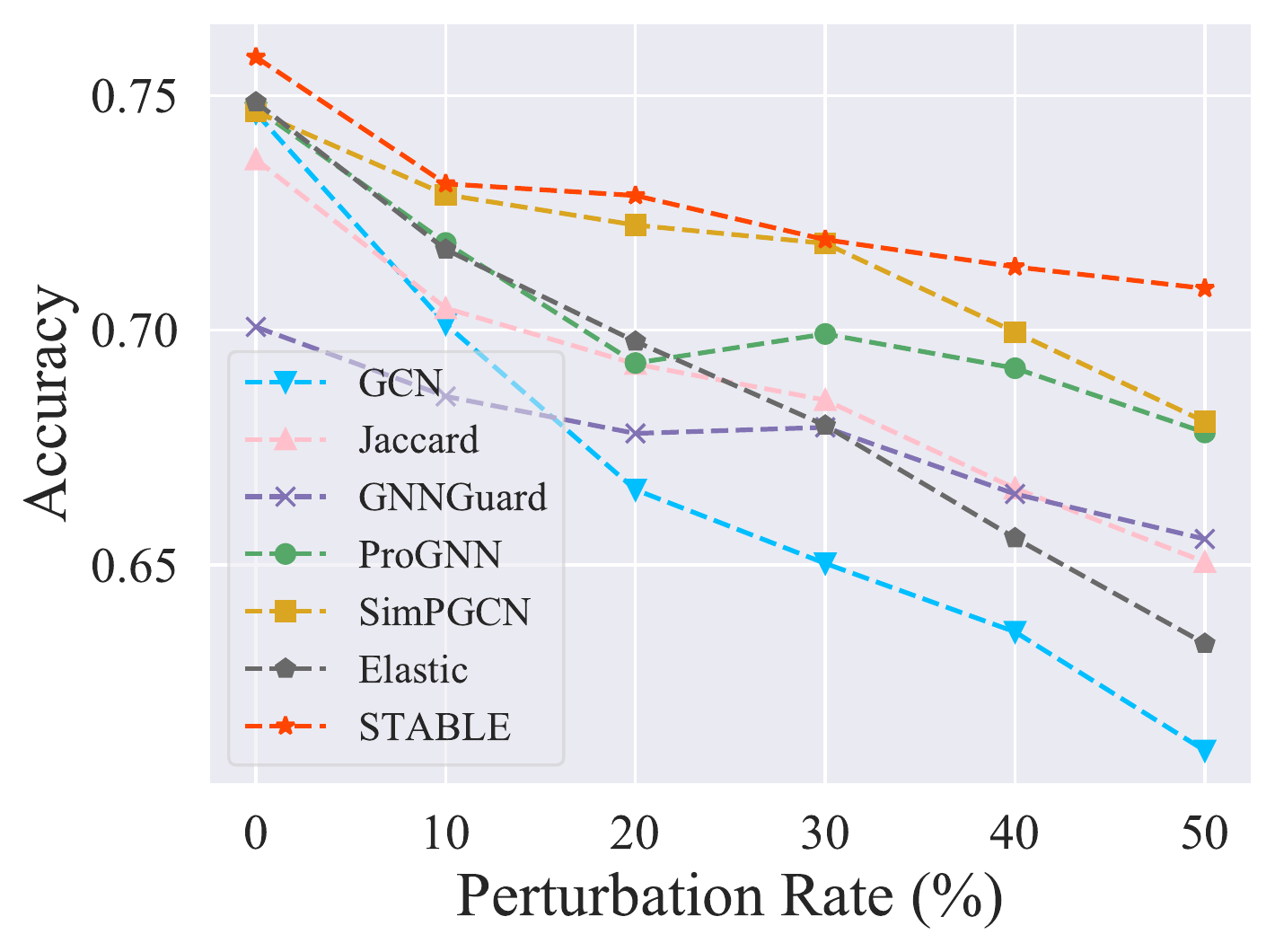}
  %\caption{fig2}
  \end{minipage}%
  }%
  \subfigure[$\textrm{Cora}_{\textrm{Random}}$]{
  \begin{minipage}[t]{0.24\textwidth}
  \centering
  \includegraphics[width=1.7in]{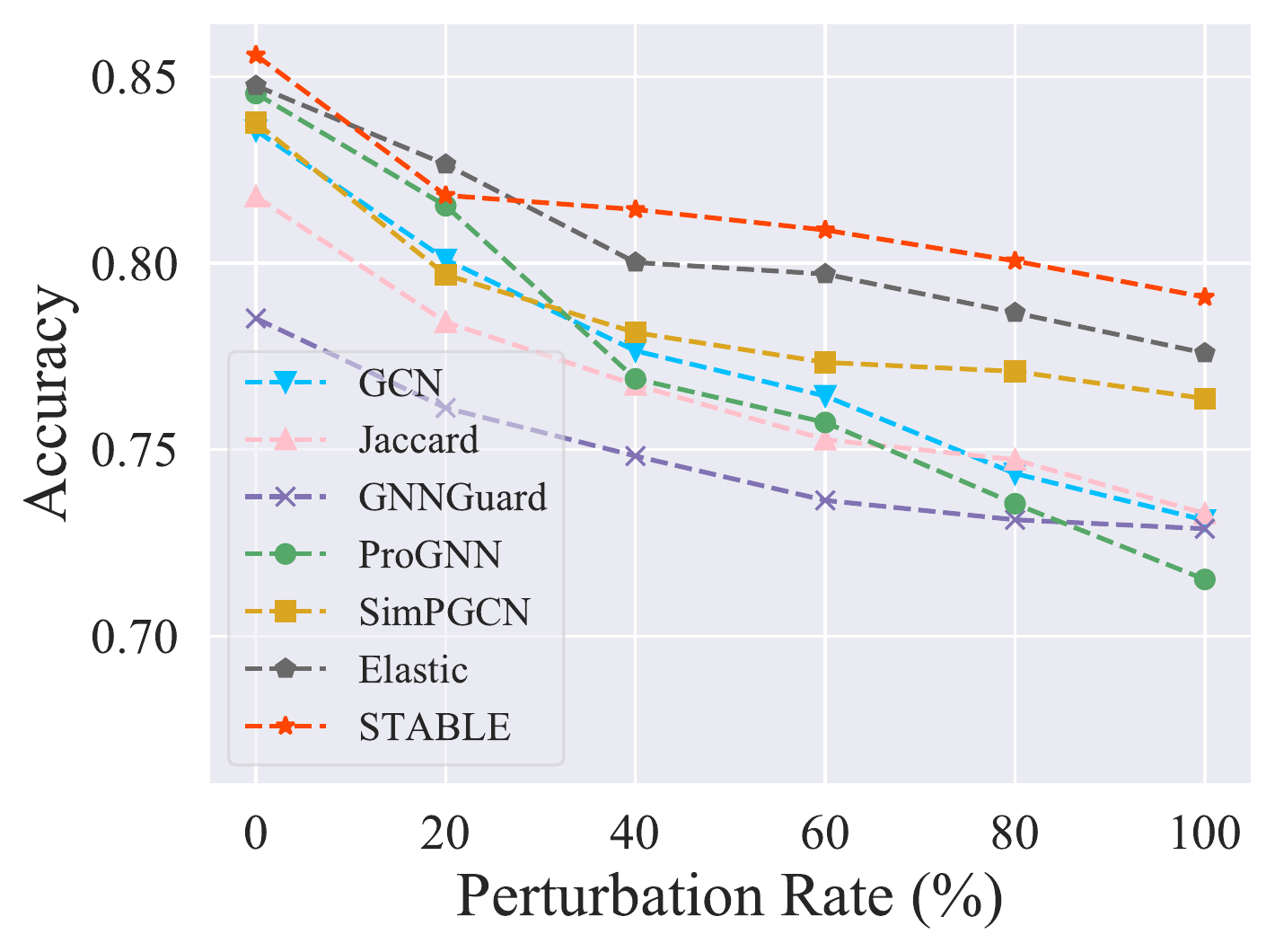}
  %\caption{fig2}
  \end{minipage}
  }%
  \subfigure[$\textrm{Citeseer}_{\textrm{Random}}$]{
  \begin{minipage}[t]{0.24\textwidth}
  \centering
  \includegraphics[width=1.7in]{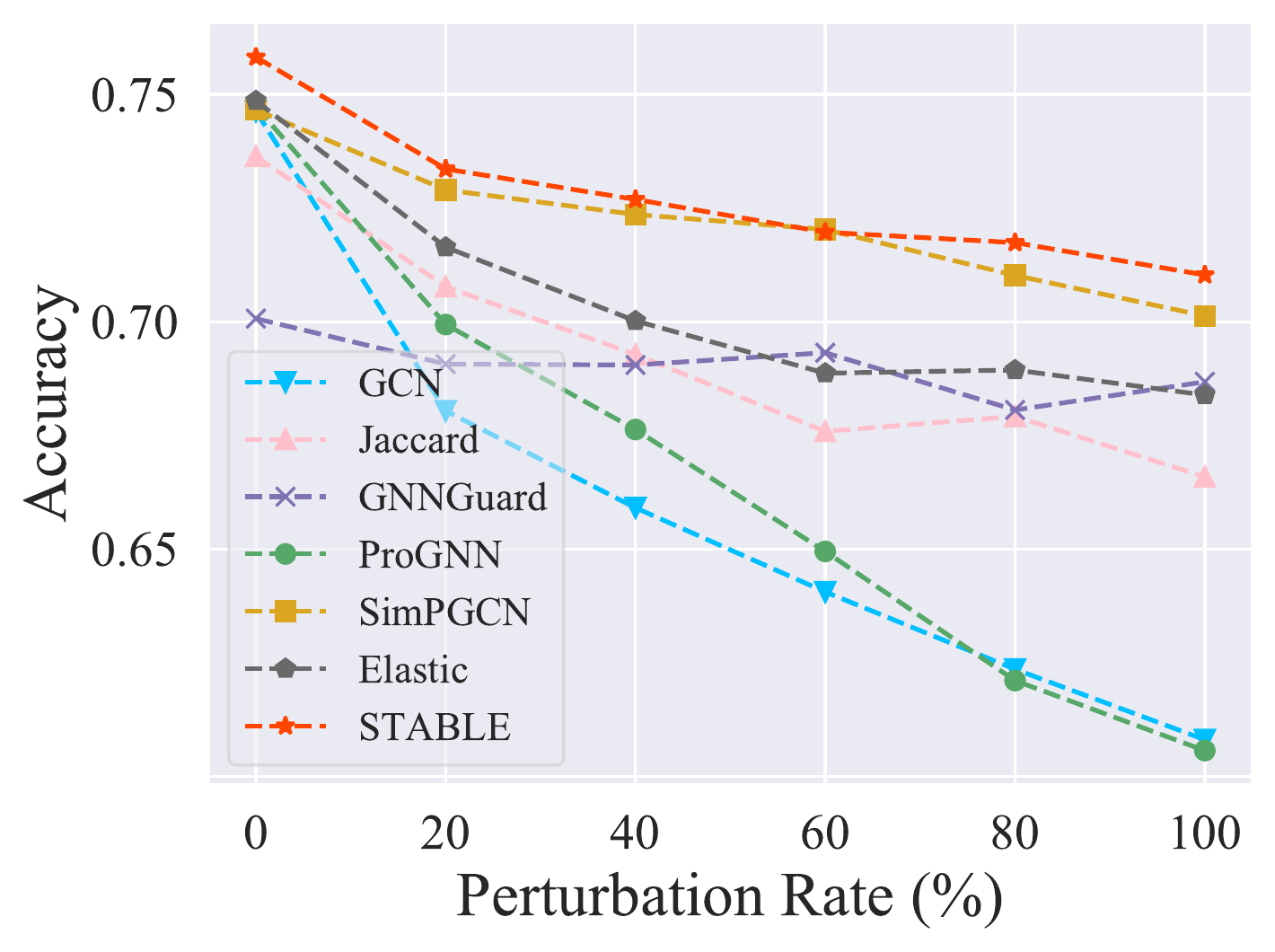}
  %\caption{fig2}
  \end{minipage}
  }%
  \centering
  \vspace{-4mm}
  \caption{Cora and Citeseer under DICE and Random}
  \label{other}
  
\end{figure*}

\subsection{Robustness Evaluation ($\mathbf{RQ1}$)}
\label{main_exp}
To answer  $\mathbf{RQ1}$, we evaluate the performance of all methods attacked by different methods.
\subsubsection{Against Mettack} Table \ref{main} shows the performance on three datasets against MetaAttack~\cite{zugner_adversarial_2019}, which is an effective attack method. The top two performance is highlighted in bold and underline. We set the perturbation rate from 0\% to 20\%. 
% Notably, existing works mostly discuss the performance under low perturbation rates to keep the attack unnoticeable. However, unlike the adversarial attack on images, people can identify at a glance whether a picture has been heavily disturbed or not~\cite{goodfellow2014generative}. Under the transductive setting, defenders have no clean data to compare, so it is hard to know whether a graph is heavily perturbated. \textbf{Therefore, we consider it valuable to study the robustness under high perturbation rates (like 35\% and 50\% in the table)}. When the perturbation rate is 50\%, one-third of the edges are adversarial edges. In other words, the graph is heavily contaminated. 
From this main experiment, we have the following observations:

\begin{itemize}[leftmargin=0.2cm]
  \item All the methods perform closely on clean graphs because most of them are designed for adversarial attacks, and the performance on clean graphs is not their goal. Both GCN and RGCN perform poorly on graphs with perturbations, proving that GNNs are vulnerable without valid defenses.
  \item Jaccard, GNNGuard, GRCN, and ProGNN are all structure learning methods. Jaccard and GNNGuard seem insensitive to the perturbations, but there is a trade-off between the performance and the robustness. They prune the edges based on the features, but the raw features cannot sufficiently represent nodes' various properties. ProGNN and GRCN perform well when the perturbation rate is low, and the accuracy drops rapidly as the perturbation rate rises. GRCN suffers from poor end-to-end representations, and for ProGNN, we guess it is hard to optimize the structure on a heavily contaminated graph directly. Compared with them, STABLE leverages the task-irrelevant representations to optimize the graph structure, which leads to higher performance.
  \item SimPGCN and Elastic are two recent state-of-the-art robust methods. SimPGCN is shown robust on Cora and Citeseer because it can adaptively balance structure and feature information. It performs poorly on Polblogs due to the lack of feature information, which can also prove that the robustness of SimpGCN relies on the raw features. Hence, it might also face the same problem that the original features miss some valuable information. Elastic benefits from its ability to model local smoothing. Different from them, our STABLE focus on refining the graph structure, and we just design a variant of GCN as the classifier. STABLE outperforms these two methods with 1\%$\thicksim$8\% improvement under low perturbation rate and 7\%$\thicksim$24\% improvement under large perturbation rate.
  \item STABLE outperforms other methods under different perturbation rates. Note that on Polblogs, we randomly remove and add a small portion of edges to generate augmentations. As the perturbation rate increases, the performance drops slowly on all datasets, which demonstrates that STABLE is insensitive to the perturbations. Especially STABLE has a huge improvement over other methods on heavily contaminated graphs.
\end{itemize}

\subsubsection{Other attacks}
The performance under DICE and Random is presented in Fig. \ref{other}. We only show the results on Cora and Citeseer due to the page limitation. RGCN and GRCN are not competitive, so they are not shown to keep the figure neat. Considering that DICE and Random are not as effective as MetaAttack, we set the perturbation rate higher. The figure shows that STABLE consistently outperforms all other baselines and successfully resists both attacks. Together with the observations from Section \ref{main_exp}, we can conclude that STABLE outperforms these baselines and is able to defend against different types of attacks.

\subsection{Result of Sturcture Learning ($\mathbf{RQ2}$)}
To validate the effectiveness of structure learning, we compare the results of structure optimization in STABLE and other metric learning  methods. RGCN fails to refine the graph and makes the graph denser, so we exclude it. The statistics of the learned graphs are shown in Table \ref{remove}. It shows the number of total removed edges, removed adversarial edges, and removed normal edges. Due to the limit of space, we only show results on Cora under MetaAttack.

To ensure a fair comparison, we tune the pruning thresholds in each method to close the number of total removed edges. It can be observed that STABLE achieves the highest pruning accuracy, indicating that STABLE revise the structure more precisely via more reliable representations.

\begin{table}[h]
%   \vspace{0.1cm}
%   \setlength{\abovecaptionskip}{0.1cm}
%   \setlength{\belowcaptionskip}{-0.4cm}

  \caption{The statistics of the removed edges on the learned graph.}
  \vspace{-2mm}
  \begin{tabular}{c|c|c|c|c}
    \toprule
    Method & Total & Adversarial & Normal & Accuracy(\%) \\
    \midrule
    Jaccard & $1,008$ & 447 & 561 & 44.35 \\
    GNNGuard & $1,082$ & 482 & 600 & 44.55 \\
    STABLE & $1,035$ & 601 & 434 & 58.07 \\
  \bottomrule
\end{tabular}  
\label{remove}
\end{table}

\subsection{Parameter Analysis ($\mathbf{RQ3}$)}

From our experimental experience, we mainly tune $k$ and $\alpha$ to achieve peak performance. Thus, we alter their values to see how they affect the robustness of STABLE. The sensitivity of other parameters are presented in Appendix \ref{t1t2 sen}. As illustrated in Fig. \ref{sensitivity}, we conduct experiments on Cora with the perturbation rate of 20\% by MetaAttack. It is worth noting that, regardless of the value of k, it is better to add than not to add. Another observation is that even $\alpha=5$, which means nodes almost only aggregate messages from the neighbors with the highest degree, the result is still better than vanilla GCN, \emph{i.e.}, $\alpha=-0.5$. 
\begin{figure}[h]
%   \vspace{-0.4cm}
  \setlength{\abovecaptionskip}{-0.1cm}
  \centering
  \subfigure{
  \begin{minipage}[t]{0.48\linewidth}
  \centering
  \includegraphics[width=1.7in]{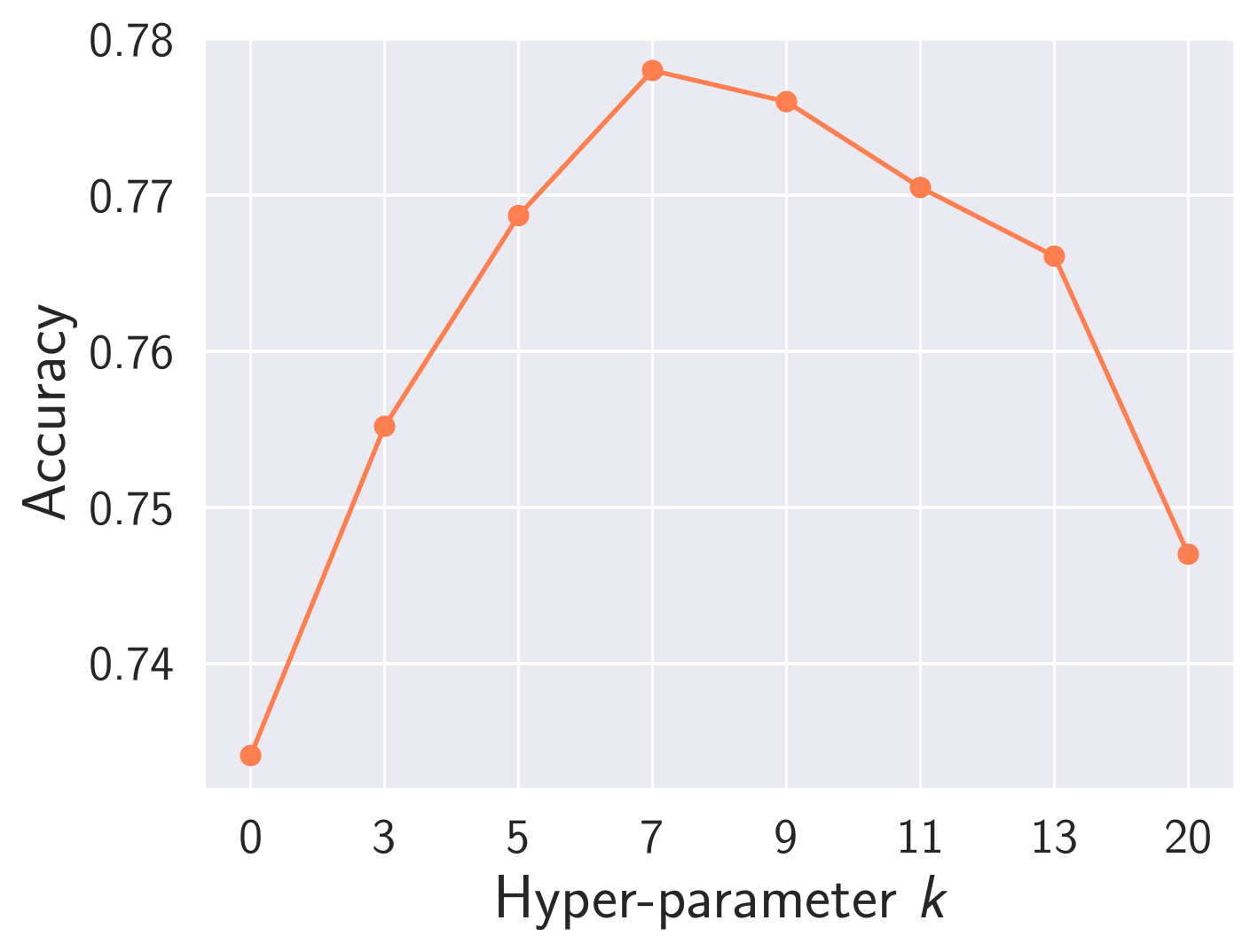}
  %\caption{fig2}
  \end{minipage}
  }%
  \subfigure{
  \begin{minipage}[t]{0.48\linewidth}
  \centering
  \includegraphics[width=1.7in]{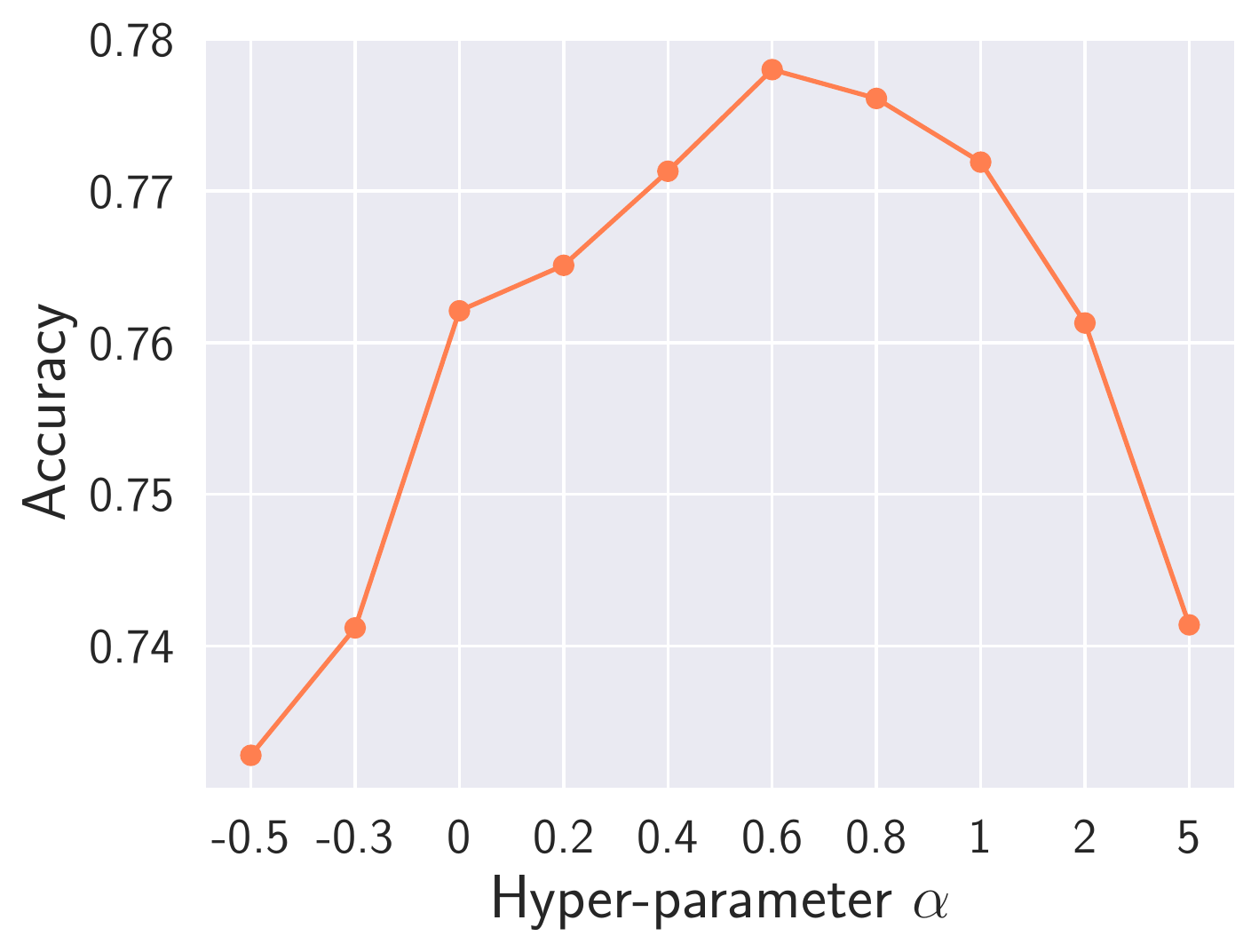}
  %\caption{fig2}
  \end{minipage}
  }%
  \centering
  \caption{Parameter sensitivity analysis on Cora.}
  \label{sensitivity}
\end{figure}

Moreover, to explore the relationship between these parameters and perturbation rates, we list the specific values which achieve the best performance on Cora in Table \ref{peak}. Both $k$ and $\alpha$ are directly proportional to the perturbation rate, which is consistent with our views. The more poisoned the graph is, the more helpful neighbors are needed, and the more trust in high-degree nodes. 

\begin{table}[h]\centering\small
%   \vspace{0.1cm}
%   \setlength{\abovecaptionskip}{0.1cm}
%   \setlength{\belowcaptionskip}{-0.4cm}
  \caption{The specific values of $k$ and $\alpha$ which achieve the peak performance on Cora under different perturbation rate}
  \vspace{-2mm}
  \begin{tabular}{c|ccccccc}
    \toprule
    Ptb Rate & 0\% & 5\% & 10\% & 15\% & 20\% & 35\% & 50\% \\
    \midrule
    $k$ & 1 & 5 & 7 & 7 & 7 & 7 & 13 \\
    $\alpha$ & -0.5 & -0.3 & 0.3 & 0.6 & 0.6 & 0.7 & 0.8 \\
  \bottomrule
\end{tabular}
\label{peak}
 \vspace{-0.2cm}
\end{table}
\section{Related Work}
In this section, we present the related literature on robust graph neural networks and graph contrastive learning.
\subsection{Robust GNNs}
Extensive studies have demonstrated that GNNs are highly fragile to adversarial attacks~\cite{dai2018adversarial,zugner2018adversarial, zugner_adversarial_2019,wu2019adversarial,zhu2022binarizedattack}. The attackers can greatly degrade the performance of GNNs by limitedly modifying the graph data, namely structure and features. 

To strengthen the robustness of GNNs, multiple methods have been proposed in the literature, including structure learning~\cite{jin2020graph}, adversarial training~\cite{xu2019topology}, utilizing Gaussian distributions to represent nodes~\cite{zhu2019robust}, designing a new message passing scheme driven by $l_1$-based graph smoothing~\cite{liu2021elastic}, combining the \emph{k}NN graph and the original graph~\cite{jin2021node}, and excluding the low-rank singular components of the graph~\cite{entezari2020all}. ~\cite{geisler2020reliable} and ~\cite{chen2021understanding} study the robustness of GNNs from the breakdown point perspective and propose more robust aggregation approaches. 

\textls[-15]{In the methods mentioned above, one representative type of approach is graph structure learning~\cite{jin2020graph,zhang2020gnnguard, luo2021learning, zhu2021deep}, which aims to detect the potential adversarial edges and assign these edges lower weights or even remove them. ProGNN~\cite{jin2020graph} and GLNN~\cite{gao2020exploring} tried to directly optimize the structure by treating it as a learnable parameter, and they introduced some regularizations like sparsity, feature smoothness, and low-rank into the objective. ~\cite{wu2019adversarial} has found that attackers tend to connect two nodes with different features, so they invented Jaccard to prune the edges that link two dissimilar nodes. GNNGuard~\cite{zhang2020gnnguard} also modeled  the edge weights by raw features. They further calculated the edge pruning  probability through a non-linear transformation. GRCN~\cite{yu2020graph} and GAUGM~\cite{zhao2021data} directly computed the weights of the edges by the inner product of representations learned by the classifier, and no additional parameters were needed.}

\textls[-18]{Different from the above-mentioned methods, we aim to leverage task-irrelevant representations, which represent rich properties of nodes and are insensitive to perturbations, to refine the graph structure. Then, we input the well-refined graph into a light robust classifier.}

\subsection{Graph Contrastive Learning}
To learn task-irrelevant representations, we consider using unsupervised methods, which are mainly divided into three categories: GAEs~\cite{kipf2016variational, garcia2017learning}, random walk methods~\cite{jeh2003scaling, perozzi2014deepwalk}, and contrastive methods~\cite{velivckovic2018deep, hassani2020contrastive, peng2020graph, Zhu:2020vf}. Both GAEs and random walk methods are suffer from the proximity over-emphasizing~\cite{velivckovic2018deep}, and the augmentation scheme in contrastive methods are naturally similar to adversarial attacks.

\textls[-15]{Graph contrastive methods are proved to be effective in node classification tasks~\cite{velivckovic2018deep, hassani2020contrastive, peng2020graph, Zhu:2020vf}. The first such method, DGI~\cite{velivckovic2018deep}, transferred the mutual information maximization~\cite{hjelm2018learning} to the graph domain. Next, InfoGraph~\cite{sun2020infograph} modified DGI’s pipeline to make the global representation useful for graph classification tasks. Recently, GRACE~\cite{Zhu:2020vf} and GraphCL~\cite{you2020graph} adapted the data augmentation methods in vision~\cite{chen2020simple} to graphs and achieved state-of-the-art performance. Specifically, GraphCL generated views by various augmentations, such as node dropping, edge perturbation, attribute masking, and subgraph. However, these augmentations were not designed for adversarial attack scenarios. The edge perturbation method in GraphCL randomly added or deleted edges in the graph, and we empirically prove this random augmentation does not work in Appendix \ref{secablation}.}

Unlike GraphCL and GRACE randomly generated augmentations, STABLE generates robustness-oriented augmentations by a novel recovery scheme, which simulates attacks to make the representations insensitive to the perturbations. 

\section{Conclusion}
To overcome the vulnerability of GNNs, we propose a novel defense model, STABLE, which successfully refines the graph structure via more reliable representations. Further, we design an advanced GCN as a downstream classifier to enhance the robustness of GCN. Our experiments demonstrate that STABLE consistently outperforms state-of-the-art baselines and can defend against different types of attacks. For future work, we aim to explore representation learning in adversarial attack scenarios. The effectiveness of STABLE proves that robust representation might be the key to GNN robustness.

\begin{acks}
This work is supported by Alibaba Group through Alibaba Innovative Research
Program. This work is supported by the National Natural Science
Foundation of China under Grant (No.61976204, U1811461, U1836206).
Xiang Ao is also supported by the Project of Youth Innovation Promotion Association CAS, Beijing Nova Program Z201100006820062. Yang Liu is also supported by China Scholarship Council. We would like to thank the anonymous reviewers for their valuable comments, and Mengda Huang, Linfeng Dong, Zidi Qin for their insightful discussions.
\end{acks}

\bibliographystyle{ACM-Reference-Format}
\bibliography{ours}


\begin{thebibliography}{55}

%%% the Web address.

\ifx \showCODEN    \undefined \def \showCODEN     #1{\unskip}     \fi
\ifx \showDOI      \undefined \def \showDOI       #1{#1}\fi
\ifx \showISBNx    \undefined \def \showISBNx     #1{\unskip}     \fi
\ifx \showISBNxiii \undefined \def \showISBNxiii  #1{\unskip}     \fi
\ifx \showISSN     \undefined \def \showISSN      #1{\unskip}     \fi
\ifx \showLCCN     \undefined \def \showLCCN      #1{\unskip}     \fi
\ifx \shownote     \undefined \def \shownote      #1{#1}          \fi
\ifx \showarticletitle \undefined \def \showarticletitle #1{#1}   \fi
\ifx \showURL      \undefined \def \showURL       {\relax}        \fi
% invisible to TeX
\providecommand\bibfield[2]{#2}
\providecommand\bibinfo[2]{#2}
\providecommand\natexlab[1]{#1}
\providecommand\showeprint[2][]{arXiv:#2}

\bibitem[\protect\citeauthoryear{Abu-El-Haija, Perozzi, Kapoor, Alipourfard,
  Lerman, Harutyunyan, Ver~Steeg, and Galstyan}{Abu-El-Haija
  et~al\mbox{.}}{2019}]%
        {abu2019mixhop}
\bibfield{author}{\bibinfo{person}{Sami Abu-El-Haija}, \bibinfo{person}{Bryan
  Perozzi}, \bibinfo{person}{Amol Kapoor}, \bibinfo{person}{Nazanin
  Alipourfard}, \bibinfo{person}{Kristina Lerman}, \bibinfo{person}{Hrayr
  Harutyunyan}, \bibinfo{person}{Greg Ver~Steeg}, {and} \bibinfo{person}{Aram
  Galstyan}.} \bibinfo{year}{2019}\natexlab{}.
\newblock \showarticletitle{Mixhop: Higher-order Graph Convolutional
  Architectures via Sparsified Neighborhood Mixing}. In
  \bibinfo{booktitle}{\emph{ICML}}. PMLR, \bibinfo{pages}{21--29}.
\newblock


\bibitem[\protect\citeauthoryear{Battaglia, Hamrick, Bapst, Sanchez-Gonzalez,
  Zambaldi, Malinowski, Tacchetti, Raposo, Santoro, Faulkner,
  et~al\mbox{.}}{Battaglia et~al\mbox{.}}{2018}]%
        {battaglia2018relational}
\bibfield{author}{\bibinfo{person}{Peter~W Battaglia},
  \bibinfo{person}{Jessica~B Hamrick}, \bibinfo{person}{Victor Bapst},
  \bibinfo{person}{Alvaro Sanchez-Gonzalez}, \bibinfo{person}{Vinicius
  Zambaldi}, \bibinfo{person}{Mateusz Malinowski}, \bibinfo{person}{Andrea
  Tacchetti}, \bibinfo{person}{David Raposo}, \bibinfo{person}{Adam Santoro},
  \bibinfo{person}{Ryan Faulkner}, {et~al\mbox{.}}}
  \bibinfo{year}{2018}\natexlab{}.
\newblock \showarticletitle{Relational Inductive Biases, Deep Learning, and
  Graph Networks}.
\newblock \bibinfo{journal}{\emph{arXiv preprint arXiv:1806.01261}}
  (\bibinfo{year}{2018}).
\newblock


\bibitem[\protect\citeauthoryear{Chen, Li, Peng, Liu, Zheng, and Yang}{Chen
  et~al\mbox{.}}{2021}]%
        {chen2021understanding}
\bibfield{author}{\bibinfo{person}{Liang Chen}, \bibinfo{person}{Jintang Li},
  \bibinfo{person}{Qibiao Peng}, \bibinfo{person}{Yang Liu},
  \bibinfo{person}{Zibin Zheng}, {and} \bibinfo{person}{Carl Yang}.}
  \bibinfo{year}{2021}\natexlab{}.
\newblock \showarticletitle{Understanding Structural Vulnerability in Graph
  Convolutional Networks}.
\newblock \bibinfo{journal}{\emph{arXiv preprint arXiv:2108.06280}}
  (\bibinfo{year}{2021}).
\newblock


\bibitem[\protect\citeauthoryear{Chen, Kornblith, Norouzi, and Hinton}{Chen
  et~al\mbox{.}}{2020a}]%
        {chen2020simple}
\bibfield{author}{\bibinfo{person}{Ting Chen}, \bibinfo{person}{Simon
  Kornblith}, \bibinfo{person}{Mohammad Norouzi}, {and}
  \bibinfo{person}{Geoffrey Hinton}.} \bibinfo{year}{2020}\natexlab{a}.
\newblock \showarticletitle{A Simple Framework for Contrastive Learning of
  Visual Representations}. In \bibinfo{booktitle}{\emph{ICML}}.
\newblock


\bibitem[\protect\citeauthoryear{Chen, Wu, and Zaki}{Chen
  et~al\mbox{.}}{2020b}]%
        {chen2020iterative}
\bibfield{author}{\bibinfo{person}{Yu Chen}, \bibinfo{person}{Lingfei Wu},
  {and} \bibinfo{person}{Mohammed Zaki}.} \bibinfo{year}{2020}\natexlab{b}.
\newblock \showarticletitle{Iterative Deep Graph Learning for Graph Neural
  Networks: Better and Robust Node Embeddings}.
\newblock \bibinfo{journal}{\emph{NeurIPS}}  \bibinfo{volume}{33}.
\newblock


\bibitem[\protect\citeauthoryear{Chien, Peng, Li, and Milenkovic}{Chien
  et~al\mbox{.}}{2021}]%
        {chien2020adaptive}
\bibfield{author}{\bibinfo{person}{Eli Chien}, \bibinfo{person}{Jianhao Peng},
  \bibinfo{person}{Pan Li}, {and} \bibinfo{person}{Olgica Milenkovic}.}
  \bibinfo{year}{2021}\natexlab{}.
\newblock \showarticletitle{Adaptive Universal Generalized Pagerank Graph
  Neural Network}. In \bibinfo{booktitle}{\emph{ICLR}}.
\newblock


\bibitem[\protect\citeauthoryear{Dai, Li, Tian, Huang, Wang, Zhu, and Song}{Dai
  et~al\mbox{.}}{2018}]%
        {dai2018adversarial}
\bibfield{author}{\bibinfo{person}{Hanjun Dai}, \bibinfo{person}{Hui Li},
  \bibinfo{person}{Tian Tian}, \bibinfo{person}{Xin Huang},
  \bibinfo{person}{Lin Wang}, \bibinfo{person}{Jun Zhu}, {and}
  \bibinfo{person}{Le Song}.} \bibinfo{year}{2018}\natexlab{}.
\newblock \showarticletitle{Adversarial Attack on Graph Structured Data}. In
  \bibinfo{booktitle}{\emph{ICML}}. \bibinfo{pages}{1115--1124}.
\newblock


\bibitem[\protect\citeauthoryear{Entezari, Al-Sayouri, Darvishzadeh, and
  Papalexakis}{Entezari et~al\mbox{.}}{2020}]%
        {entezari2020all}
\bibfield{author}{\bibinfo{person}{Negin Entezari}, \bibinfo{person}{Saba~A
  Al-Sayouri}, \bibinfo{person}{Amirali Darvishzadeh}, {and}
  \bibinfo{person}{Evangelos~E Papalexakis}.} \bibinfo{year}{2020}\natexlab{}.
\newblock \showarticletitle{All You Need is Low (rank) Defending Against
  Adversarial Attacks on Graphs}. In \bibinfo{booktitle}{\emph{WSDM}}.
  \bibinfo{pages}{169--177}.
\newblock


\bibitem[\protect\citeauthoryear{Fatemi, Asri, and Kazemi}{Fatemi
  et~al\mbox{.}}{2021}]%
        {fatemi2021slaps}
\bibfield{author}{\bibinfo{person}{Bahare Fatemi}, \bibinfo{person}{Layla~El
  Asri}, {and} \bibinfo{person}{Seyed~Mehran Kazemi}.}
  \bibinfo{year}{2021}\natexlab{}.
\newblock \showarticletitle{SLAPS: Self-Supervision Improves Structure Learning
  for Graph Neural Networks}.
\newblock


\bibitem[\protect\citeauthoryear{Finkelshtein, Baskin, Zheltonozhskii, and
  Alon}{Finkelshtein et~al\mbox{.}}{2020}]%
        {finkelshtein2020single}
\bibfield{author}{\bibinfo{person}{Ben Finkelshtein}, \bibinfo{person}{Chaim
  Baskin}, \bibinfo{person}{Evgenii Zheltonozhskii}, {and} \bibinfo{person}{Uri
  Alon}.} \bibinfo{year}{2020}\natexlab{}.
\newblock \showarticletitle{Single-Node Attack for Fooling Graph Neural
  Networks}.
\newblock \bibinfo{journal}{\emph{arXiv preprint arXiv:2011.03574}}
  (\bibinfo{year}{2020}).
\newblock


\bibitem[\protect\citeauthoryear{Gao, Hu, and Guo}{Gao et~al\mbox{.}}{2020}]%
        {gao2020exploring}
\bibfield{author}{\bibinfo{person}{Xiang Gao}, \bibinfo{person}{Wei Hu}, {and}
  \bibinfo{person}{Zongming Guo}.} \bibinfo{year}{2020}\natexlab{}.
\newblock \showarticletitle{Exploring Structure-adaptive Graph Learning for
  Robust Semi-supervised Classification}. In \bibinfo{booktitle}{\emph{ICME}}.
  IEEE.
\newblock


\bibitem[\protect\citeauthoryear{Garcia~Duran and Niepert}{Garcia~Duran and
  Niepert}{2017}]%
        {garcia2017learning}
\bibfield{author}{\bibinfo{person}{Alberto Garcia~Duran} {and}
  \bibinfo{person}{Mathias Niepert}.} \bibinfo{year}{2017}\natexlab{}.
\newblock \showarticletitle{Learning Graph Representations With Embedding
  Propagation}. In \bibinfo{booktitle}{\emph{NeurIPS}},
  Vol.~\bibinfo{volume}{30}.
\newblock


\bibitem[\protect\citeauthoryear{Geisler, Z{\"u}gner, and
  G{\"u}nnemann}{Geisler et~al\mbox{.}}{2020}]%
        {geisler2020reliable}
\bibfield{author}{\bibinfo{person}{Simon Geisler}, \bibinfo{person}{Daniel
  Z{\"u}gner}, {and} \bibinfo{person}{Stephan G{\"u}nnemann}.}
  \bibinfo{year}{2020}\natexlab{}.
\newblock \showarticletitle{Reliable Graph Neural Networks via Robust
  Aggregation}. In \bibinfo{booktitle}{\emph{NeurIPS}}.
\newblock


\bibitem[\protect\citeauthoryear{Hamilton, Ying, and Leskovec}{Hamilton
  et~al\mbox{.}}{2017}]%
        {hamilton2017inductive}
\bibfield{author}{\bibinfo{person}{William~L Hamilton}, \bibinfo{person}{Rex
  Ying}, {and} \bibinfo{person}{Jure Leskovec}.}
  \bibinfo{year}{2017}\natexlab{}.
\newblock \showarticletitle{Inductive Representation Learning on Large Graphs}.
  In \bibinfo{booktitle}{\emph{NeruIPS}}. \bibinfo{pages}{1025--1035}.
\newblock


\bibitem[\protect\citeauthoryear{Hassani and Khasahmadi}{Hassani and
  Khasahmadi}{2020}]%
        {hassani2020contrastive}
\bibfield{author}{\bibinfo{person}{Kaveh Hassani} {and}
  \bibinfo{person}{Amir~Hosein Khasahmadi}.} \bibinfo{year}{2020}\natexlab{}.
\newblock \showarticletitle{Contrastive Multi-view Representation Learning on
  Graphs}. In \bibinfo{booktitle}{\emph{ICML}}. PMLR,
  \bibinfo{pages}{4116--4126}.
\newblock


\bibitem[\protect\citeauthoryear{Hjelm, Fedorov, Lavoie-Marchildon, Grewal,
  Bachman, Trischler, and Bengio}{Hjelm et~al\mbox{.}}{2019}]%
        {hjelm2018learning}
\bibfield{author}{\bibinfo{person}{R~Devon Hjelm}, \bibinfo{person}{Alex
  Fedorov}, \bibinfo{person}{Samuel Lavoie-Marchildon}, \bibinfo{person}{Karan
  Grewal}, \bibinfo{person}{Phil Bachman}, \bibinfo{person}{Adam Trischler},
  {and} \bibinfo{person}{Yoshua Bengio}.} \bibinfo{year}{2019}\natexlab{}.
\newblock \showarticletitle{Learning Deep Representations by Mutual Information
  Estimation and Maximization}. In \bibinfo{booktitle}{\emph{ICLR}}.
\newblock


\bibitem[\protect\citeauthoryear{Hou, Zhang, Cheng, Ma, Ma, Chen, and Yang}{Hou
  et~al\mbox{.}}{2019}]%
        {hou2019measuring}
\bibfield{author}{\bibinfo{person}{Yifan Hou}, \bibinfo{person}{Jian Zhang},
  \bibinfo{person}{James Cheng}, \bibinfo{person}{Kaili Ma},
  \bibinfo{person}{Richard~TB Ma}, \bibinfo{person}{Hongzhi Chen}, {and}
  \bibinfo{person}{Ming-Chang Yang}.} \bibinfo{year}{2019}\natexlab{}.
\newblock \showarticletitle{Measuring and Improving The Use of Graph
  Information in Graph Neural Networks}. In \bibinfo{booktitle}{\emph{ICLR}}.
\newblock


\bibitem[\protect\citeauthoryear{Huang, Liu, Ao, Li, Chi, Feng, Yang, and
  He}{Huang et~al\mbox{.}}{2022}]%
        {huang2022auc}
\bibfield{author}{\bibinfo{person}{Mengda Huang}, \bibinfo{person}{Yang Liu},
  \bibinfo{person}{Xiang Ao}, \bibinfo{person}{Kuan Li},
  \bibinfo{person}{Jianfeng Chi}, \bibinfo{person}{Jinghua Feng},
  \bibinfo{person}{Hao Yang}, {and} \bibinfo{person}{Qing He}.}
  \bibinfo{year}{2022}\natexlab{}.
\newblock \showarticletitle{AUC-oriented Graph Neural Network for Fraud
  Detection}. In \bibinfo{booktitle}{\emph{WWW}}.
\newblock


\bibitem[\protect\citeauthoryear{Jeh and Widom}{Jeh and Widom}{2003}]%
        {jeh2003scaling}
\bibfield{author}{\bibinfo{person}{Glen Jeh} {and} \bibinfo{person}{Jennifer
  Widom}.} \bibinfo{year}{2003}\natexlab{}.
\newblock \showarticletitle{Scaling Personalized Web Search}. In
  \bibinfo{booktitle}{\emph{WWW}}.
\newblock


\bibitem[\protect\citeauthoryear{Jin, Derr, Wang, Ma, Liu, and Tang}{Jin
  et~al\mbox{.}}{2021}]%
        {jin2021node}
\bibfield{author}{\bibinfo{person}{Wei Jin}, \bibinfo{person}{Tyler Derr},
  \bibinfo{person}{Yiqi Wang}, \bibinfo{person}{Yao Ma}, \bibinfo{person}{Zitao
  Liu}, {and} \bibinfo{person}{Jiliang Tang}.} \bibinfo{year}{2021}\natexlab{}.
\newblock \showarticletitle{Node Similarity Preserving Graph Convolutional
  Networks}. In \bibinfo{booktitle}{\emph{WSDM}}. \bibinfo{pages}{148--156}.
\newblock


\bibitem[\protect\citeauthoryear{Jin, Li, Xu, Wang, Ji, Aggarwal, and Tang}{Jin
  et~al\mbox{.}}{2020a}]%
        {jin2020adversarial}
\bibfield{author}{\bibinfo{person}{Wei Jin}, \bibinfo{person}{Yaxin Li},
  \bibinfo{person}{Han Xu}, \bibinfo{person}{Yiqi Wang},
  \bibinfo{person}{Shuiwang Ji}, \bibinfo{person}{Charu Aggarwal}, {and}
  \bibinfo{person}{Jiliang Tang}.} \bibinfo{year}{2020}\natexlab{a}.
\newblock \showarticletitle{Adversarial Attacks and Defenses on Graphs: A
  Review, A Tool and Empirical Studies}. In \bibinfo{booktitle}{\emph{KDD
  Explorations}}.
\newblock


\bibitem[\protect\citeauthoryear{Jin, Ma, Liu, Tang, Wang, and Tang}{Jin
  et~al\mbox{.}}{2020b}]%
        {jin2020graph}
\bibfield{author}{\bibinfo{person}{Wei Jin}, \bibinfo{person}{Yao Ma},
  \bibinfo{person}{Xiaorui Liu}, \bibinfo{person}{Xianfeng Tang},
  \bibinfo{person}{Suhang Wang}, {and} \bibinfo{person}{Jiliang Tang}.}
  \bibinfo{year}{2020}\natexlab{b}.
\newblock \showarticletitle{Graph Structure Learning for Robust Graph Neural
  Networks}. In \bibinfo{booktitle}{\emph{KDD}}.
\newblock


\bibitem[\protect\citeauthoryear{Kipf and Welling}{Kipf and Welling}{2016}]%
        {kipf2016variational}
\bibfield{author}{\bibinfo{person}{Thomas~N Kipf} {and} \bibinfo{person}{Max
  Welling}.} \bibinfo{year}{2016}\natexlab{}.
\newblock \showarticletitle{Variational Graph Auto-Encoders}.
\newblock \bibinfo{journal}{\emph{NIPS Workshop on Bayesian Deep Learning}}
  (\bibinfo{year}{2016}).
\newblock


\bibitem[\protect\citeauthoryear{Kipf and Welling}{Kipf and Welling}{2017}]%
        {kipf2017semi}
\bibfield{author}{\bibinfo{person}{Thomas~N. Kipf} {and} \bibinfo{person}{Max
  Welling}.} \bibinfo{year}{2017}\natexlab{}.
\newblock \showarticletitle{Semi-Supervised Classification with Graph
  Convolutional Networks}. In \bibinfo{booktitle}{\emph{ICLR}}.
\newblock


\bibitem[\protect\citeauthoryear{Li, Wang, Zhu, and Huang}{Li
  et~al\mbox{.}}{2018}]%
        {li2018adaptive}
\bibfield{author}{\bibinfo{person}{Ruoyu Li}, \bibinfo{person}{Sheng Wang},
  \bibinfo{person}{Feiyun Zhu}, {and} \bibinfo{person}{Junzhou Huang}.}
  \bibinfo{year}{2018}\natexlab{}.
\newblock \showarticletitle{Adaptive Graph Convolutional Neural Networks}. In
  \bibinfo{booktitle}{\emph{AAAI}}, Vol.~\bibinfo{volume}{32}.
\newblock


\bibitem[\protect\citeauthoryear{Li, Jin, Xu, and Tang}{Li
  et~al\mbox{.}}{2020}]%
        {li2020deeprobust}
\bibfield{author}{\bibinfo{person}{Yaxin Li}, \bibinfo{person}{Wei Jin},
  \bibinfo{person}{Han Xu}, {and} \bibinfo{person}{Jiliang Tang}.}
  \bibinfo{year}{2020}\natexlab{}.
\newblock \showarticletitle{Deeprobust: A Pytorch Library for Adversarial
  Attacks and Defenses}.
\newblock \bibinfo{journal}{\emph{arXiv preprint arXiv:2005.06149}}
  (\bibinfo{year}{2020}).
\newblock


\bibitem[\protect\citeauthoryear{Li, Tarlow, Brockschmidt, and Zemel}{Li
  et~al\mbox{.}}{2016}]%
        {li2015gated}
\bibfield{author}{\bibinfo{person}{Yujia Li}, \bibinfo{person}{Daniel Tarlow},
  \bibinfo{person}{Marc Brockschmidt}, {and} \bibinfo{person}{Richard Zemel}.}
  \bibinfo{year}{2016}\natexlab{}.
\newblock \showarticletitle{Gated Graph Sequence Neural Networks}.
\newblock \bibinfo{journal}{\emph{ICLR}}.
\newblock


\bibitem[\protect\citeauthoryear{Liu, Jin, Ma, Li, Liu, Wang, Yan, and
  Tang}{Liu et~al\mbox{.}}{2021b}]%
        {liu2021elastic}
\bibfield{author}{\bibinfo{person}{Xiaorui Liu}, \bibinfo{person}{Wei Jin},
  \bibinfo{person}{Yao Ma}, \bibinfo{person}{Yaxin Li}, \bibinfo{person}{Hua
  Liu}, \bibinfo{person}{Yiqi Wang}, \bibinfo{person}{Ming Yan}, {and}
  \bibinfo{person}{Jiliang Tang}.} \bibinfo{year}{2021}\natexlab{b}.
\newblock \showarticletitle{Elastic Graph Neural Networks}. In
  \bibinfo{booktitle}{\emph{ICML}}. PMLR, \bibinfo{pages}{6837--6849}.
\newblock


\bibitem[\protect\citeauthoryear{Liu, Ao, Qin, Chi, Feng, Yang, and He}{Liu
  et~al\mbox{.}}{2021a}]%
        {liu2021pick}
\bibfield{author}{\bibinfo{person}{Yang Liu}, \bibinfo{person}{Xiang Ao},
  \bibinfo{person}{Zidi Qin}, \bibinfo{person}{Jianfeng Chi},
  \bibinfo{person}{Jinghua Feng}, \bibinfo{person}{Hao Yang}, {and}
  \bibinfo{person}{Qing He}.} \bibinfo{year}{2021}\natexlab{a}.
\newblock \showarticletitle{Pick and Choose: A GNN-based Imbalanced Learning
  Approach for Fraud Detection}. In \bibinfo{booktitle}{\emph{WWW}}.
  \bibinfo{pages}{3168--3177}.
\newblock


\bibitem[\protect\citeauthoryear{Luo, Cheng, Yu, Zong, Ni, Chen, and Zhang}{Luo
  et~al\mbox{.}}{2021}]%
        {luo2021learning}
\bibfield{author}{\bibinfo{person}{Dongsheng Luo}, \bibinfo{person}{Wei Cheng},
  \bibinfo{person}{Wenchao Yu}, \bibinfo{person}{Bo Zong},
  \bibinfo{person}{Jingchao Ni}, \bibinfo{person}{Haifeng Chen}, {and}
  \bibinfo{person}{Xiang Zhang}.} \bibinfo{year}{2021}\natexlab{}.
\newblock \showarticletitle{Learning to Drop: Robust Graph Neural Network via
  Topological Denoising}. In \bibinfo{booktitle}{\emph{WSDM}}.
  \bibinfo{pages}{779--787}.
\newblock


\bibitem[\protect\citeauthoryear{McPherson, Smith-Lovin, and Cook}{McPherson
  et~al\mbox{.}}{2001}]%
        {mcpherson2001birds}
\bibfield{author}{\bibinfo{person}{Miller McPherson}, \bibinfo{person}{Lynn
  Smith-Lovin}, {and} \bibinfo{person}{James~M Cook}.}
  \bibinfo{year}{2001}\natexlab{}.
\newblock \showarticletitle{Birds of a Feather: Homophily in Social Networks}.
\newblock \bibinfo{journal}{\emph{Annual review of sociology}}
  \bibinfo{volume}{27}, \bibinfo{number}{1} (\bibinfo{year}{2001}),
  \bibinfo{pages}{415--444}.
\newblock


\bibitem[\protect\citeauthoryear{Peng, Huang, Luo, Zheng, Rong, Xu, and
  Huang}{Peng et~al\mbox{.}}{2020}]%
        {peng2020graph}
\bibfield{author}{\bibinfo{person}{Zhen Peng}, \bibinfo{person}{Wenbing Huang},
  \bibinfo{person}{Minnan Luo}, \bibinfo{person}{Qinghua Zheng},
  \bibinfo{person}{Yu Rong}, \bibinfo{person}{Tingyang Xu}, {and}
  \bibinfo{person}{Junzhou Huang}.} \bibinfo{year}{2020}\natexlab{}.
\newblock \showarticletitle{Graph Representation Learning via Graphical Mutual
  Information Maximization}. In \bibinfo{booktitle}{\emph{WWW}}.
\newblock


\bibitem[\protect\citeauthoryear{Perozzi, Al-Rfou, and Skiena}{Perozzi
  et~al\mbox{.}}{2014}]%
        {perozzi2014deepwalk}
\bibfield{author}{\bibinfo{person}{Bryan Perozzi}, \bibinfo{person}{Rami
  Al-Rfou}, {and} \bibinfo{person}{Steven Skiena}.}
  \bibinfo{year}{2014}\natexlab{}.
\newblock \showarticletitle{Deepwalk: Online Learning of Social
  Representations}. In \bibinfo{booktitle}{\emph{KDD}}.
  \bibinfo{pages}{701--710}.
\newblock


\bibitem[\protect\citeauthoryear{Sun, Hoffmann, Verma, and Tang}{Sun
  et~al\mbox{.}}{2020}]%
        {sun2020infograph}
\bibfield{author}{\bibinfo{person}{Fan-Yun Sun}, \bibinfo{person}{Jordan
  Hoffmann}, \bibinfo{person}{Vikas Verma}, {and} \bibinfo{person}{Jian Tang}.}
  \bibinfo{year}{2020}\natexlab{}.
\newblock \showarticletitle{Infograph: Unsupervised and Semi-supervised
  Graph-level Representation Learning via Mutual Information maximization}. In
  \bibinfo{booktitle}{\emph{ICLR}}.
\newblock


\bibitem[\protect\citeauthoryear{Veli{\v{c}}kovi{\'c}, Cucurull, Casanova,
  Romero, Lio, and Bengio}{Veli{\v{c}}kovi{\'c} et~al\mbox{.}}{2017}]%
        {velivckovic2017graph}
\bibfield{author}{\bibinfo{person}{Petar Veli{\v{c}}kovi{\'c}},
  \bibinfo{person}{Guillem Cucurull}, \bibinfo{person}{Arantxa Casanova},
  \bibinfo{person}{Adriana Romero}, \bibinfo{person}{Pietro Lio}, {and}
  \bibinfo{person}{Yoshua Bengio}.} \bibinfo{year}{2017}\natexlab{}.
\newblock \showarticletitle{Graph Attention Networks}. In
  \bibinfo{booktitle}{\emph{ICLR}}.
\newblock


\bibitem[\protect\citeauthoryear{Veli{\v{c}}kovi{\'c}, Fedus, Hamilton,
  Li{\`o}, Bengio, and Hjelm}{Veli{\v{c}}kovi{\'c} et~al\mbox{.}}{2018}]%
        {velivckovic2018deep}
\bibfield{author}{\bibinfo{person}{Petar Veli{\v{c}}kovi{\'c}},
  \bibinfo{person}{William Fedus}, \bibinfo{person}{William~L Hamilton},
  \bibinfo{person}{Pietro Li{\`o}}, \bibinfo{person}{Yoshua Bengio}, {and}
  \bibinfo{person}{R~Devon Hjelm}.} \bibinfo{year}{2018}\natexlab{}.
\newblock \showarticletitle{Deep Graph Infomax}.
\newblock \bibinfo{journal}{\emph{arXiv preprint arXiv:1809.10341}}
  (\bibinfo{year}{2018}).
\newblock


\bibitem[\protect\citeauthoryear{Wang, Zhu, Bo, Cui, Shi, and Pei}{Wang
  et~al\mbox{.}}{2020}]%
        {wang2020gcn}
\bibfield{author}{\bibinfo{person}{Xiao Wang}, \bibinfo{person}{Meiqi Zhu},
  \bibinfo{person}{Deyu Bo}, \bibinfo{person}{Peng Cui}, \bibinfo{person}{Chuan
  Shi}, {and} \bibinfo{person}{Jian Pei}.} \bibinfo{year}{2020}\natexlab{}.
\newblock \showarticletitle{Am-gcn: Adaptive Multi-channel Graph Convolutional
  Networks}. In \bibinfo{booktitle}{\emph{KDD}}. \bibinfo{pages}{1243--1253}.
\newblock


\bibitem[\protect\citeauthoryear{Waniek, Michalak, Wooldridge, and
  Rahwan}{Waniek et~al\mbox{.}}{2018}]%
        {waniek2018hiding}
\bibfield{author}{\bibinfo{person}{Marcin Waniek}, \bibinfo{person}{Tomasz~P
  Michalak}, \bibinfo{person}{Michael~J Wooldridge}, {and}
  \bibinfo{person}{Talal Rahwan}.} \bibinfo{year}{2018}\natexlab{}.
\newblock \showarticletitle{Hiding Individuals and Communities in A Social
  Network}.
\newblock \bibinfo{journal}{\emph{Nature Human Behaviour}} \bibinfo{volume}{2},
  \bibinfo{number}{2} (\bibinfo{year}{2018}), \bibinfo{pages}{139--147}.
\newblock


\bibitem[\protect\citeauthoryear{Wiles, Gowal, Stimberg, Alvise-Rebuffi, Ktena,
  Cemgil, et~al\mbox{.}}{Wiles et~al\mbox{.}}{2022}]%
        {wiles2021fine}
\bibfield{author}{\bibinfo{person}{Olivia Wiles}, \bibinfo{person}{Sven Gowal},
  \bibinfo{person}{Florian Stimberg}, \bibinfo{person}{Sylvestre
  Alvise-Rebuffi}, \bibinfo{person}{Ira Ktena}, \bibinfo{person}{Taylan
  Cemgil}, {et~al\mbox{.}}} \bibinfo{year}{2022}\natexlab{}.
\newblock \showarticletitle{A Fine-grained Analysis on Distribution Shift}. In
  \bibinfo{booktitle}{\emph{ICLR}}.
\newblock


\bibitem[\protect\citeauthoryear{Wu, Wang, Tyshetskiy, Docherty, Lu, and
  Zhu}{Wu et~al\mbox{.}}{2019}]%
        {wu2019adversarial}
\bibfield{author}{\bibinfo{person}{Huijun Wu}, \bibinfo{person}{Chen Wang},
  \bibinfo{person}{Yuriy Tyshetskiy}, \bibinfo{person}{Andrew Docherty},
  \bibinfo{person}{Kai Lu}, {and} \bibinfo{person}{Liming Zhu}.}
  \bibinfo{year}{2019}\natexlab{}.
\newblock \showarticletitle{Adversarial Examples on Graph Data: Deep Insights
  Into Attack and Defense}.
\newblock \bibinfo{journal}{\emph{IJCAI}}.
\newblock


\bibitem[\protect\citeauthoryear{Xu, Chen, Liu, Chen, Weng, Hong, and Lin}{Xu
  et~al\mbox{.}}{2019a}]%
        {xu2019topology}
\bibfield{author}{\bibinfo{person}{Kaidi Xu}, \bibinfo{person}{Hongge Chen},
  \bibinfo{person}{Sijia Liu}, \bibinfo{person}{Pin-Yu Chen},
  \bibinfo{person}{Tsui-Wei Weng}, \bibinfo{person}{Mingyi Hong}, {and}
  \bibinfo{person}{Xue Lin}.} \bibinfo{year}{2019}\natexlab{a}.
\newblock \showarticletitle{Topology Attack and Defense for Graph Neural
  Networks: An Optimization Perspective}.
\newblock \bibinfo{journal}{\emph{IJCAI}}.
\newblock


\bibitem[\protect\citeauthoryear{Xu, Hu, Leskovec, and Jegelka}{Xu
  et~al\mbox{.}}{2019b}]%
        {xu2019powerful}
\bibfield{author}{\bibinfo{person}{Keyulu Xu}, \bibinfo{person}{Weihua Hu},
  \bibinfo{person}{Jure Leskovec}, {and} \bibinfo{person}{Stefanie Jegelka}.}
  \bibinfo{year}{2019}\natexlab{b}.
\newblock \showarticletitle{How Powerful Are Graph Neural Networks?}. In
  \bibinfo{booktitle}{\emph{ICLR}}.
\newblock


\bibitem[\protect\citeauthoryear{You, Chen, Sui, Chen, Wang, and Shen}{You
  et~al\mbox{.}}{2020}]%
        {you2020graph}
\bibfield{author}{\bibinfo{person}{Yuning You}, \bibinfo{person}{Tianlong
  Chen}, \bibinfo{person}{Yongduo Sui}, \bibinfo{person}{Ting Chen},
  \bibinfo{person}{Zhangyang Wang}, {and} \bibinfo{person}{Yang Shen}.}
  \bibinfo{year}{2020}\natexlab{}.
\newblock \showarticletitle{Graph Contrastive Learning With Augmentations}.
\newblock \bibinfo{journal}{\emph{NeurIPS}}.
\newblock


\bibitem[\protect\citeauthoryear{Yu, Zhang, Jiang, Wu, and Yang}{Yu
  et~al\mbox{.}}{2020}]%
        {yu2020graph}
\bibfield{author}{\bibinfo{person}{Donghan Yu}, \bibinfo{person}{Ruohong
  Zhang}, \bibinfo{person}{Zhengbao Jiang}, \bibinfo{person}{Yuexin Wu}, {and}
  \bibinfo{person}{Yiming Yang}.} \bibinfo{year}{2020}\natexlab{}.
\newblock \showarticletitle{Graph-revised Convolutional Network}. In
  \bibinfo{booktitle}{\emph{ECML}}. Springer, \bibinfo{pages}{378--393}.
\newblock


\bibitem[\protect\citeauthoryear{Zhang and Zitnik}{Zhang and Zitnik}{2020}]%
        {zhang2020gnnguard}
\bibfield{author}{\bibinfo{person}{Xiang Zhang} {and} \bibinfo{person}{Marinka
  Zitnik}.} \bibinfo{year}{2020}\natexlab{}.
\newblock \showarticletitle{GNNGuard: Defending Graph Neural Networks Against
  Adversarial Attacks}.
\newblock \bibinfo{journal}{\emph{NeurIPS}}.
\newblock


\bibitem[\protect\citeauthoryear{Zhang, Cui, and Zhu}{Zhang
  et~al\mbox{.}}{2020}]%
        {zhang2020deep}
\bibfield{author}{\bibinfo{person}{Ziwei Zhang}, \bibinfo{person}{Peng Cui},
  {and} \bibinfo{person}{Wenwu Zhu}.} \bibinfo{year}{2020}\natexlab{}.
\newblock \showarticletitle{Deep learning on graphs: A survey}.
\newblock \bibinfo{journal}{\emph{IEEE Transactions on Knowledge and Data
  Engineering}} (\bibinfo{year}{2020}).
\newblock


\bibitem[\protect\citeauthoryear{Zhao, Liu, Neves, Woodford, Jiang, and
  Shah}{Zhao et~al\mbox{.}}{2021}]%
        {zhao2021data}
\bibfield{author}{\bibinfo{person}{Tong Zhao}, \bibinfo{person}{Yozen Liu},
  \bibinfo{person}{Leonardo Neves}, \bibinfo{person}{Oliver Woodford},
  \bibinfo{person}{Meng Jiang}, {and} \bibinfo{person}{Neil Shah}.}
  \bibinfo{year}{2021}\natexlab{}.
\newblock \showarticletitle{Data Augmentation for Graph Neural Networks}. In
  \bibinfo{booktitle}{\emph{AAAI}}.
\newblock


\bibitem[\protect\citeauthoryear{Zhou, Cui, Hu, Zhang, Yang, Liu, Wang, Li, and
  Sun}{Zhou et~al\mbox{.}}{2020}]%
        {zhou2020graph}
\bibfield{author}{\bibinfo{person}{Jie Zhou}, \bibinfo{person}{Ganqu Cui},
  \bibinfo{person}{Shengding Hu}, \bibinfo{person}{Zhengyan Zhang},
  \bibinfo{person}{Cheng Yang}, \bibinfo{person}{Zhiyuan Liu},
  \bibinfo{person}{Lifeng Wang}, \bibinfo{person}{Changcheng Li}, {and}
  \bibinfo{person}{Maosong Sun}.} \bibinfo{year}{2020}\natexlab{}.
\newblock \showarticletitle{Graph Neural Networks: A Review of Methods and
  Applications}.
\newblock \bibinfo{journal}{\emph{AI Open}}  \bibinfo{volume}{1}
  (\bibinfo{year}{2020}), \bibinfo{pages}{57--81}.
\newblock


\bibitem[\protect\citeauthoryear{Zhu, Zhang, Cui, and Zhu}{Zhu
  et~al\mbox{.}}{2019}]%
        {zhu2019robust}
\bibfield{author}{\bibinfo{person}{Dingyuan Zhu}, \bibinfo{person}{Ziwei
  Zhang}, \bibinfo{person}{Peng Cui}, {and} \bibinfo{person}{Wenwu Zhu}.}
  \bibinfo{year}{2019}\natexlab{}.
\newblock \showarticletitle{Robust Graph Convolutional Networks Against
  Adversarial Attacks}. In \bibinfo{booktitle}{\emph{KDD}}.
  \bibinfo{pages}{1399--1407}.
\newblock


\bibitem[\protect\citeauthoryear{Zhu, Ao, Qin, Chang, Liu, He, and Li}{Zhu
  et~al\mbox{.}}{2021a}]%
        {zhu2021intelligent}
\bibfield{author}{\bibinfo{person}{Xiaoqian Zhu}, \bibinfo{person}{Xiang Ao},
  \bibinfo{person}{Zidi Qin}, \bibinfo{person}{Yanpeng Chang},
  \bibinfo{person}{Yang Liu}, \bibinfo{person}{Qing He}, {and}
  \bibinfo{person}{Jianping Li}.} \bibinfo{year}{2021}\natexlab{a}.
\newblock \showarticletitle{Intelligent financial fraud detection practices in
  post-pandemic era}.
\newblock \bibinfo{journal}{\emph{The Innovation}} \bibinfo{volume}{2},
  \bibinfo{number}{4} (\bibinfo{year}{2021}), \bibinfo{pages}{100176}.
\newblock


\bibitem[\protect\citeauthoryear{Zhu, Lai, Zhao, Luo, Yuan, Ren, and Zhou}{Zhu
  et~al\mbox{.}}{2022}]%
        {zhu2022binarizedattack}
\bibfield{author}{\bibinfo{person}{Yulin Zhu}, \bibinfo{person}{Yuni Lai},
  \bibinfo{person}{Kaifa Zhao}, \bibinfo{person}{Xiapu Luo},
  \bibinfo{person}{Mingquan Yuan}, \bibinfo{person}{Jian Ren}, {and}
  \bibinfo{person}{Kai Zhou}.} \bibinfo{year}{2022}\natexlab{}.
\newblock \showarticletitle{BinarizedAttack: Structural Poisoning Attacks to
  Graph-based Anomaly Detection}. In \bibinfo{booktitle}{\emph{ICDE}}.
\newblock


\bibitem[\protect\citeauthoryear{Zhu, Xu, Zhang, Liu, Wu, and Wang}{Zhu
  et~al\mbox{.}}{2021b}]%
        {zhu2021deep}
\bibfield{author}{\bibinfo{person}{Yanqiao Zhu}, \bibinfo{person}{Weizhi Xu},
  \bibinfo{person}{Jinghao Zhang}, \bibinfo{person}{Qiang Liu},
  \bibinfo{person}{Shu Wu}, {and} \bibinfo{person}{Liang Wang}.}
  \bibinfo{year}{2021}\natexlab{b}.
\newblock \showarticletitle{Deep Graph Structure Learning for Robust
  Representations: A Survey}.
\newblock \bibinfo{journal}{\emph{arXiv preprint arXiv:2103.03036}}.
\newblock


\bibitem[\protect\citeauthoryear{Zhu, Xu, Yu, Liu, Wu, and Wang}{Zhu
  et~al\mbox{.}}{2020}]%
        {Zhu:2020vf}
\bibfield{author}{\bibinfo{person}{Yanqiao Zhu}, \bibinfo{person}{Yichen Xu},
  \bibinfo{person}{Feng Yu}, \bibinfo{person}{Qiang Liu}, \bibinfo{person}{Shu
  Wu}, {and} \bibinfo{person}{Liang Wang}.} \bibinfo{year}{2020}\natexlab{}.
\newblock \showarticletitle{{Deep Graph Contrastive Representation Learning}}.
  In \bibinfo{booktitle}{\emph{ICML Workshop on Graph Representation Learning
  and Beyond}}.
\newblock
\urldef\tempurl%
\url{http://arxiv.org/abs/2006.04131}
\showURL{%
\tempurl}


\bibitem[\protect\citeauthoryear{Z{\"u}gner, Akbarnejad, and
  G{\"u}nnemann}{Z{\"u}gner et~al\mbox{.}}{2018}]%
        {zugner2018adversarial}
\bibfield{author}{\bibinfo{person}{Daniel Z{\"u}gner}, \bibinfo{person}{Amir
  Akbarnejad}, {and} \bibinfo{person}{Stephan G{\"u}nnemann}.}
  \bibinfo{year}{2018}\natexlab{}.
\newblock \showarticletitle{Adversarial Attacks on Neural Networks for Graph
  Data}. In \bibinfo{booktitle}{\emph{KDD}}. \bibinfo{pages}{2847--2856}.
\newblock


\bibitem[\protect\citeauthoryear{Z{\"u}gner and G{\"u}nnemann}{Z{\"u}gner and
  G{\"u}nnemann}{2019}]%
        {zugner_adversarial_2019}
\bibfield{author}{\bibinfo{person}{Daniel Z{\"u}gner} {and}
  \bibinfo{person}{Stephan G{\"u}nnemann}.} \bibinfo{year}{2019}\natexlab{}.
\newblock \showarticletitle{Adversarial Attacks on Graph Neural Networks via
  Meta Learning}. In \bibinfo{booktitle}{\emph{ICLR}}.
\newblock


\end{thebibliography}

\appendix

\section{APPENDIX}
\subsection{Algorithm}
\label{algorithm}

The overall training algorithm is shown in Algorithm \ref{algori}. In line 2, we roughly pre-process $\mathcal{G}$ by removing some potential perturbations. From lines 3 to 10, we utilize a contrastive learning model with robustness-oriented augmentations to obtain the node representations. In lines 11 and 12, we refine the graph structure based on the representations learned before. from line 13 to 20, we train the classifier $\textrm{GCN}^*$ on $G^*$.
\begin{algorithm}[h]
  \caption{STABLE}
  \label{algori}
  \SetAlgoLined
  \KwIn{Graph $\mathcal{G}=\{\mathcal{V}, \mathcal{E}, \mathcal{X}\}$, Labels $\mathcal{Y}_L$, pre-process threshold $t_1$, recovery probability $p$, pruning threshold $t_2$, number of adding helpful neighbors $k$, advanced GCN parameters $\alpha$ and $\beta$, training epochs $N_{epoch}$}%输入参数
  \KwOut{The predicted labels of $\mathcal{V}_U$}%输出
  \BlankLine

  Initialize parameters $\phi$, $\omega$ and $\theta$\;
  Roughly pre-process $\mathcal{G}$ to obtain $\mathcal{G}^P$ using Eq. (\ref{pre-process}) and $t_1$\;
  Generate augmentations $\mathcal{G}^P_1$ to $\mathcal{G}^P_M$ using Eq. (\ref{augmentation})\;
  Generate $\widetilde{\mathcal{G}}^P$ by shuffling the node features of $\mathcal{G}^P$\;
  \While{not converge}{
  Compute $\textbf{H}$, $\widetilde{\textbf{H}}$, $\textbf{H}_1$, $\textbf{H}_2$..., and $\textbf{H}_M$ using Eq. (\ref{encoder})\;
  Compute the global representations of $\mathcal{G}^P_1$ to $\mathcal{G}^P_M$ using Eq. (\ref{readout})\;
  Compute the contrastive learning objective using Eq.(\ref{lossc})\;
  Update $\phi$ and $\omega$ by $\frac{\partial \mathcal{L}_C}{\partial \phi}$ and $\frac{\partial \mathcal{L}_C}{\partial \omega}$\;
  }
  Prune the strcutrue using Eq. (\ref{similarity}) and Eq.(\ref{prune})\;
  Add helpful edges using Eq. (\ref{addedge})\;
  \For{e=1, ..., $N_{epoch}$}{
  \For{t=1, 2}{
  \For{$v_i \in \mathcal{V}$}{
  Compute $h_i$ using Eq. (\ref{prop})\;
  }
  }
  Compute $\mathcal{L}_G$ using Eq. (\ref{lossg})\;
  Update $\theta$ by $\frac{\partial \mathcal{L}_G}{\partial \theta}$
  }
  return argmax($f_\theta (\textbf{H}, \mathbf{A}^*)_{\mathcal{V}_U}$, dim=1)\; 
  \end{algorithm}

\subsection{Datasets}
\label{dataset}
Following ~\cite{zugner2018adversarial, jin2020graph}, we only consider the largest connected connected component (LCC). The statistics is listed in Tabel \ref{Data statistics}. There is no features available in Polblogs. Following ~\cite{jin2020graph, liu2021elastic} we set the feature matrix to be a $n\times n$ identity matrix. The results of GNNGuard and Jaccard on Polblogs are not available because the cosine similarity of the identity matrix is meaningless. For PubMed dataset, we use the attacked graphs provided provided by~\cite{jin2020adversarial}.
\begin{table}[h]
  \caption{Dataset statistics. We only consider the largest connected component\ (LCC).}
  \label{Data statistics}
  \begin{tabular}{c|cccc}
    \toprule
    Datasets & $\textrm{N}_{\textrm{LCC}}$ & $\textrm{E}_{\textrm{LCC}}$ & Classes & Features \\
    \midrule
    Cora & 2,485 &5,069 & 7 & 1433 \\
    Citeseer & 2,110 &3,668 & 6 & 3703 \\
    Polblogs & 1,222 &16,714 & 2 & / \\
    PubMed &19717 & 44338 & 3 & 500\\
  \bottomrule
\end{tabular}
\end{table}

\subsection{Baselines}
\label{baselines intro}
\begin{itemize}
  \item \textbf{GCN}~\cite{kipf2017semi}: GCN is a popular graph convolutional network based
  on spectral theory. 
  \item \textbf{RGCN}~\cite{zhu2019robust}: RGCN utilizes gaussian distributions to represent node and uses a variance-based attention mechanism to remedy the propagation of adversarial
  attacks.
  \item \textbf{Jaccard}~\cite{wu2019adversarial}: Since attacks tend to link nodes with different labels, Jaccard prune edges which connect two dissimilar nodes.
  \item \textbf{GNNGuard}~\cite{zhang2020gnnguard}: GNNGuard utilizes cosine similarity to model the edge weights and then calculates edge pruning probability through a non-linear transformation.
  \item \textbf{GRCN}~\cite{yu2020graph}: GRCN models edge weights by taking inner product of embeddings of two end nodes.
  \item \textbf{ProGNN}~\cite{jin2020graph}: ProGNN treats the adjacency matrix as learnable parameters and directly optimizes it with three regularizations, \emph{i.e.}, feature smoothness, low-rank and sparsity.
  \item \textbf{SimpGCN}~\cite{jin2021node}: SimpGCN utilizes a $k$NN graph to keep the nodes with similar features close in the representation space and a self-learning regularization to keep the nodes with dissimilar features remote.
  \item \textbf{Elastic}~\cite{liu2021elastic}: Elastic introduces $\ell_1$-norm to graph signal estimator and proposes elastic message passing which is derived from one step optimization of such estimator. The local smoothness adaptivity enables the Elastic GNNs robust to structural attacks.
  \item \textbf{MetaAttack}~\cite{zugner_adversarial_2019}: MetaAttack uses meta-gradients to solve the bilevel problem underlying poisoning attacks, essentially treating the graph as a hyperparameter to optimize.
  \item \textbf{DICE}~\cite{waniek2018hiding}: Disconnect Internally, connect externally.
  \item \textbf{Random}: Inject random structure noise.
\end{itemize}

\subsection{Implementation Details}
\label{implementation}
We use DeepRobust, an adversarial attack repository~\cite{li2020deeprobust}, to implement all the attack methods, RGCN, ProGNN, SimpGCN, and Jaccard. GNNGuard, Elastic, and GCN are implemented with the code provided by the authors.

For each graph, we randomly split the nodes into 10\% for training, 10\% for validation, and 80\% for testing. Then we generate attacks on each graph according to the perturbation rate, and all the hyper-parameters in attack methods are the same with the authors' implementation. For all the defense models, we report the average accuracy and standard deviation of 10 runs. 

All the hyper-parameters are tuned based on the loss and accuracy of the validation set. For RGCN, the hidden dimensions are tuned from \{16, 32, 64, 128\}. For Jaccard, the Jaccard Similarity threshold are tuned from \{0.01, 0.02, 0.03, 0.04, 0.05\}. For GNNGuard, we follow the author's settings, which contains $P_0=0.5$, $K=2$, $D_2=16$ and $dropout=0.5$. For ProGNN and SimpGCN, following~\cite{jin2021node}, we use the default hyper-parameter settings in the authors’ implementation. For Elastic, the propagation step K is tuned from \{3, 5, 10\}, and the parameter $\lambda_1$ and $\lambda_2$ are tuned from \{0, 3, 6, 9\}.

For our work, $t_1$ is Jaccard Similarity threshold in this work and tuned from\{0.0, 0.01, 0.02, 0.03, 0.04, 0.05\}, the recovery portion $p$ is fixed at 0.2, $t_2$ is tuned from \{0.1, 0.2, 0.3\}, $k$ is tuned from \{1, 3, 5, 7, 11, 13\}, $\alpha$ is tuned from $-0.5$ to $3$, and $\beta$ is fixed at 2. We fix $M=2$ because we find that two augmentation views are good enough. The other parameters in $GCN^*$ follows the setting in ~\cite{kipf2017semi}.

\subsection{Ablation Study ($\mathbf{RQ4}$)}
\label{secablation}
We conduct an ablation study to examine the contributions of different components in STABLE:

\begin{itemize}[leftmargin=0.2cm]
  \item \textbf{STABLE-P}: STABLE without rough pre-process.
  \item \textbf{STABLE-A}: STABLE without augmentations.
  \item \textbf{STABLE-Ran}: STABLE with random augmentations. We generate the views by randomly removing or adding edges.
  %\item \textbf{STABLE-R}: STABLE without representation learning part. The graph is revised by raw features.
  \item \textbf{STABLE-K}: We only prune edges in the refining phase.
  \item \textbf{STABLE-GCN}: Replace advanced GCN with GCN.
\end{itemize}
\begin{figure}[h]
%   \vspace{-0.4cm}
%   \setlength{\belowcaptionskip}{-0.4cm}
  \centering
  \subfigure[Cora]{
  \begin{minipage}[t]{0.48\linewidth}
  \centering
  \includegraphics[width=1.7in]{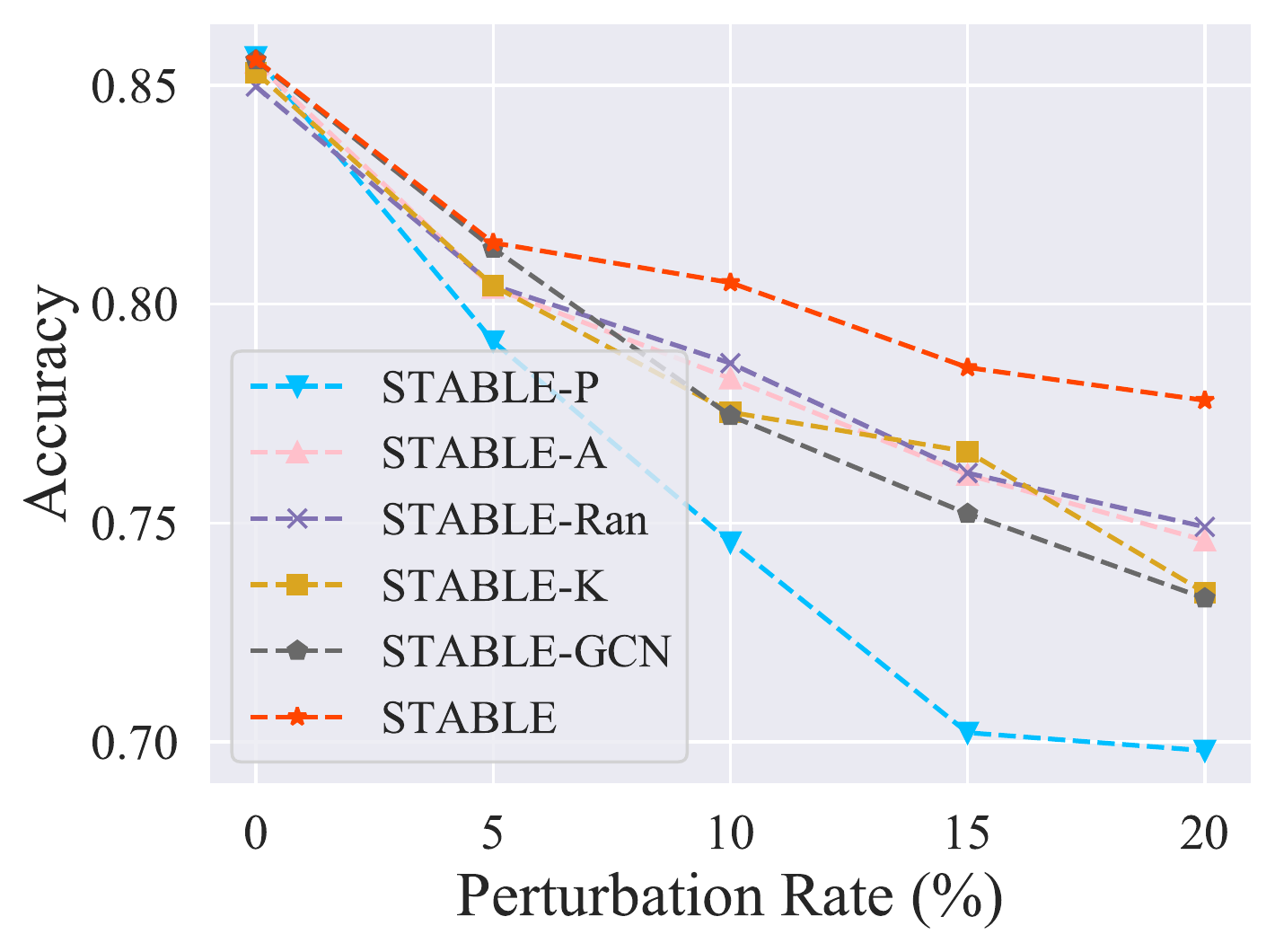}
  %\caption{fig2}
  \end{minipage}
  }%
  \subfigure[Citeseer]{
  \begin{minipage}[t]{0.48\linewidth}
  \centering
  \includegraphics[width=1.7in]{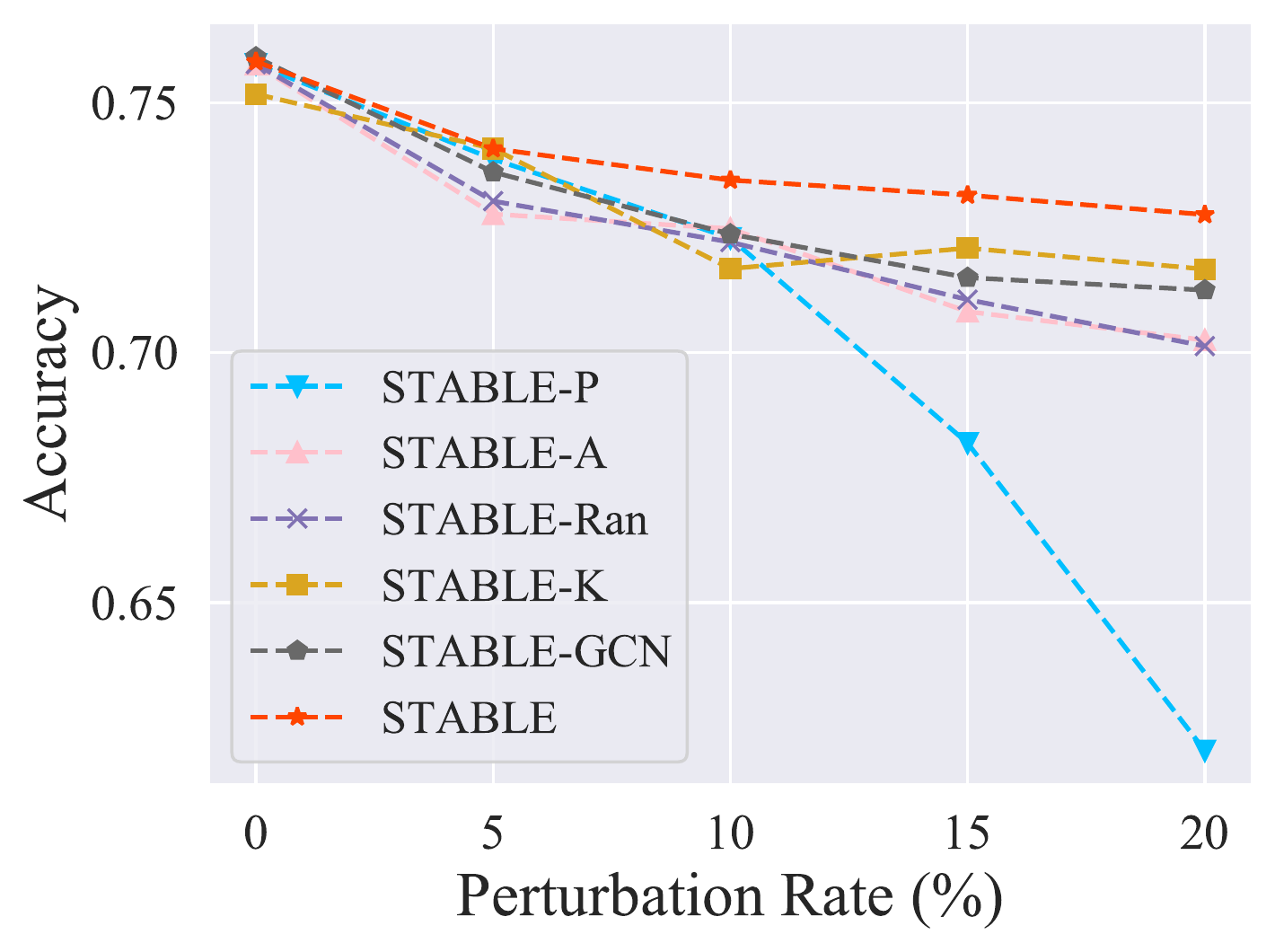}
  %\caption{fig2}
  \end{minipage}
  }%
  \centering
  \vspace{-4mm}
  \caption{Comparisons between STABLE and its variants.}
  \label{ablation}
\end{figure}

The accuracy on Cora and Citeseer under MetaAttack is illustrated in Fig. \ref{ablation}. It is clearly observed that the STABLE-P perform worst, and it is in line with our point that $\mathcal{G}^P$ is much cleaner than $\mathcal{G}$. STABLE-Ran and STABLE-A perform very closely, and the gap between them and STABLE widens as the perturbation rate increases. We can conclude that the robustness-oriented augmentation indeed works. It is worth noting that STABLE outperforms STABLE-K more significantly as the perturbation rate rising, as our expectation. We argue that adding helpful neighbors can diminish the impact of harmful edges, especially on heavily poisoned graphs. 
% STABLE-R refines the graph by raw features, so it performs like Jaccard and GNNGuard. In general, it is robust but performs poorly under low perturbation rate.

To further verify the robustness of advanced GCN, we present the performance of SimpGCN* and Jaccard* in Table \ref{advanced performance}. Jaccard and SimpGCN both contain the vanilla GCN, so we can easily define two variants by replacing the GCN with the advanced GCN. As shown in Table \ref{advanced performance}, the variants of Jaccard and SimpGCN show more robust than the original models. The improvement of SimpGCN is more significant might  because of the pruning strategy in Jaccard, which already removes some adversarial edges.
\begin{table}[h]\centering\small
  \caption{Classification accuracy(\%) on Cora under different perturbation rates. The asterisk indicates that the GCN part of this model is replaced with advanced GCN.}
  \begin{tabular}{c|c|cccc}
    \toprule
    Datasets & Ptb rate & Jaccard & Jaccard* & SimpGCN & SimpGCN* \\
    \midrule
    \multirow{7}*{Cora} & 0\% & \textbf{81.79} & 81.11 & \textbf{83.77} & 83.64 \\
                        & 5\% & 80.23 & \textbf{80.57} & 78.98 & \textbf{80.45} \\
                        & 10\% & 74.65 & \textbf{76.99} & 75.07 & \textbf{78.04} \\
                        & 15\% & 74.29 & \textbf{76.32} & 71.42 & \textbf{75.31} \\
                        & 20\% & 73.11 & \textbf{73.42} & 68.90 & \textbf{73.29} \\
                        & 35\% & 66.11 & \textbf{68.79} & 64.87 & \textbf{71.15} \\
                        & 50\% & 58.08 & \textbf{64.06} & 51.94 &\textbf{65.63} \\
  \bottomrule
\end{tabular}
\label{advanced performance}
\end{table}

\subsection{Sensitivity}
\label{t1t2 sen}
\begin{figure}[h]
%   \vspace{-0.4cm}
  \setlength{\abovecaptionskip}{-0.1cm}
  \centering
  \subfigure{
  \begin{minipage}[t]{0.48\linewidth}
  \centering
  \includegraphics[width=1.7in]{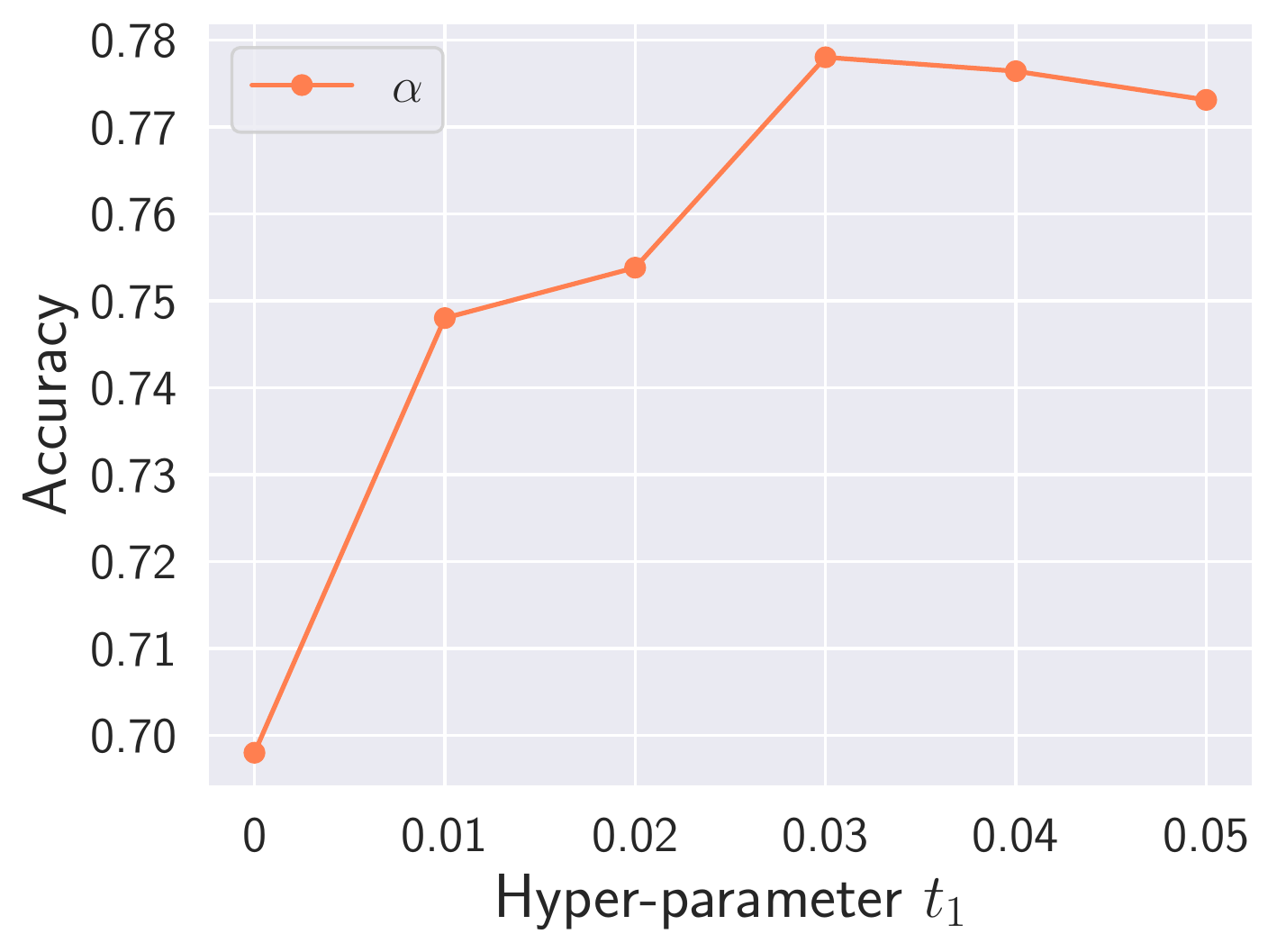}
  %\caption{fig2}
  \end{minipage}
  }%
  \subfigure{
  \begin{minipage}[t]{0.48\linewidth}
  \centering
  \includegraphics[width=1.7in]{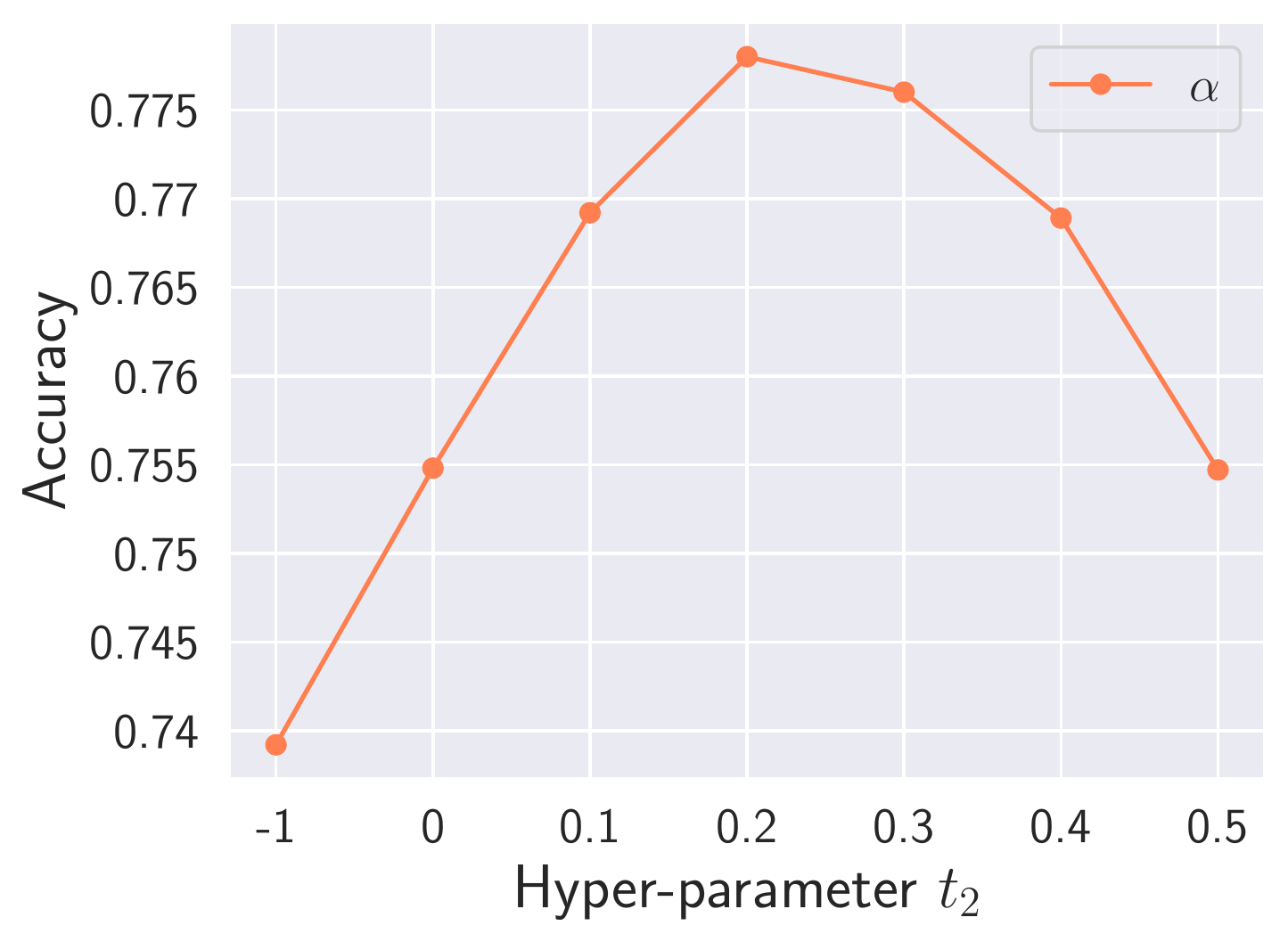}
  %\caption{fig2}
  \end{minipage}
  }%
  \centering
  \caption{Parameter sensitivity analysis on Cora.}
  \label{sensitivity_t}
\end{figure}

We explore the sensitivity of $t_1$ and $t_2$ for STABLE. The performance change of STABLE is illustrate in Figure \ref{sensitivity_t}. In fact, for Cora, we fixed $t_1$ at 0.03 and $t_2$ at 0.2 to achieve the best performance under different perturbation rate. For $t_1$, it is worth noting that, the performance on pre-processed graph is consistently higher than on the original graph($t_1=0$). $t_2=-1$ implies that no edge is pruned in the refining step, and the performance is still competitive might due to the good representations and helpful edge insertions.

\end{document}